\documentclass{article} % For LaTeX2e
\usepackage{template/iclr2024_conference,times}

\usepackage{hyperref}
\usepackage{url}
\usepackage{outlines}
\usepackage{amsmath}
\usepackage{amsfonts}
\usepackage{graphicx}

\usepackage{booktabs}
\usepackage{multirow}
\usepackage{rotating}
\usepackage{wrapfig}
\usepackage{pifont}
\title{Fast Hyperboloid Decision Tree Algorithms}

\author{Philippe Chlenski,$^1$ Ethan Turok,$^1$ Antonio Moretti,$^2$ Itsik Pe'er$^{1}$\\
$^1$Columbia University $^2$Barnard College\\
}

\newcommand{\hyperdt}{\textsc{HyperDT}}
\newcommand{\hyperrf}{\textsc{HyperRF}}
\newcommand{\hororf}{\textsc{HoroRF}}
\newcommand{\horodt}{\textsc{HoroDT}}

\newcommand{\x}{\mathbf{x}}
\newcommand{\nd}{\mathbf{n}(d)}
\newcommand{\ndt}{\mathbf{n}(d, \theta)}

\newcommand{\sklrn}{\textsc{scikit-learn}}
\newcommand{\hdk}{\mathbb{H}^{D,K}}
\iclrfinalcopy % Uncomment for camera-ready version, but NOT for submission.
\begin{document}

\maketitle

\begin{abstract}
% Hyperbolic geometry is gaining traction in machine learning for its effectiveness at capturing hierarchical structures in real-world data. Hyperbolic spaces, where neighborhoods grow exponentially, offer substantial advantages and consistently deliver state-of-the-art results across diverse applications. However, hyperbolic classifiers often grapple with computational challenges. Methods reliant on Riemannian optimization frequently exhibit sluggishness, stemming from the increased computational demands of operations on Riemannian manifolds. In response to these challenges, we present \hyperdt, a novel extension of decision tree algorithms into hyperbolic space. Crucially, \hyperdt\ eliminates the need for computationally intensive Riemannian optimization, numerically unstable exponential and logarithmic maps, or pairwise comparisons between points by leveraging inner products to adapt Euclidean decision tree algorithms to hyperbolic space. Our approach is conceptually straightforward and maintains constant-time decision complexity while mitigating the scalability issues inherent in high-dimensional Euclidean spaces. Building upon \hyperdt\ we introduce \hyperrf, a hyperbolic random forest model. Extensive benchmarking across diverse datasets underscores the superior performance of these models, providing a swift, precise, accurate, and user-friendly toolkit for hyperbolic data analysis.
Hyperbolic geometry is gaining traction in machine learning due to its capacity to effectively capture hierarchical structures in real-world data. Hyperbolic spaces,
where neighborhoods grow exponentially, offer substantial advantages and have consistently delivered state-of-the-art results across diverse applications. However, hyperbolic classifiers often grapple with computational challenges. Methods reliant on Riemannian optimization frequently exhibit sluggishness, stemming from the increased computational demands of operations on Riemannian manifolds. In response to these challenges, we present \hyperdt, a novel extension of decision tree algorithms into hyperbolic space. Crucially, \hyperdt eliminates the need for computationally intensive Riemannian optimization, numerically unstable exponential and logarithmic maps, or pairwise comparisons between points by leveraging inner products to adapt Euclidean decision tree algorithms to hyperbolic space. Our approach is conceptually straightforward and maintains constant-time decision complexity while mitigating the scalability issues inherent in high-dimensional Euclidean spaces. Building upon \hyperdt\ we introduce \hyperrf, a hyperbolic random forest model. Extensive benchmarking across diverse datasets underscores the superior performance of these models, providing a swift, precise, accurate, and user-friendly toolkit for hyperbolic data analysis. Our code can be found at~\url{https://github.com/pchlenski/hyperdt}.
\end{abstract}

\section{Introduction}
\subsection{Background: Hyperbolic Embeddings}
The adoption of hyperbolic geometry for graph embeddings has sparked a vibrant and rapidly growing body of machine learning research~\citep{van_kreveld_low_2012,chamberlain_neural_2017,gu_learning_2019,chami_trees_2020,chami_horopca_2021}. This surge in interest is driven by the compelling advantages offered by hyperbolic spaces, particularly in capturing hierarchical and tree-like structures inherent in various real-world datasets. In hyperbolic space, neighborhoods grow exponentially rather than polynomially, allowing for embeddings of exponentially growing systems such as phylogenetic trees or concept hierarchies. Hyperbolic embeddings have proven to be highly effective, showcasing state-of-the-art results across various applications including question answering \citep{tay_hyperbolic_2018}, node classification~\citep{chami_trees_2020},  and word embeddings \citep{tifrea_poincare_2018}. These achievements underscore the growing interest in classifiers that operate natively within hyperbolic spaces~\citep{gulcehre_hyperbolic_2018, marconi_hyperbolic_2020, doorenbos_hyperbolic_2023}. %These classifiers leverage the unique curvature properties of hyperbolic geometry to make more nuanced and accurate predictions of hierarchical data.

%While hyperbolic classifiers leverage the unique curvature properties of hyperbolic geometry to make more nuanced and accurate predictions of hierarchical data, such techniques often face a dilemma. Some methods are consistent with this geometry but suffer from time-complexity penalties associated with horosphere calculations (see, for example, ~\citep{fan_horospherical_2023, doorenbos_hyperbolic_2023}). Others abandon negative curvature when conducting inference and resort to Euclidean predictors, transforming data into Euclidean representations through logarithmic maps and employing whitening techniques~\citep{chami_horopca_2021} or directly applying Euclidean predictors to hyperbolic data. This introduces an additional layer of complexity to the inference process, which can have adverse effects on both speed and interpretability (citation needed). Additionally, methods employing Riemannian optimization often exhibit sluggishness due to the increased computational complexity associated with operations on Riemannian manifolds, which require intricate geometric calculations. The sensitivity of Riemannian optimization to initialization and the presence of complex geometric constraints can further contribute to slower convergence.

\begin{figure}[ht]
    \centering
    \includegraphics[width=0.8\textwidth]{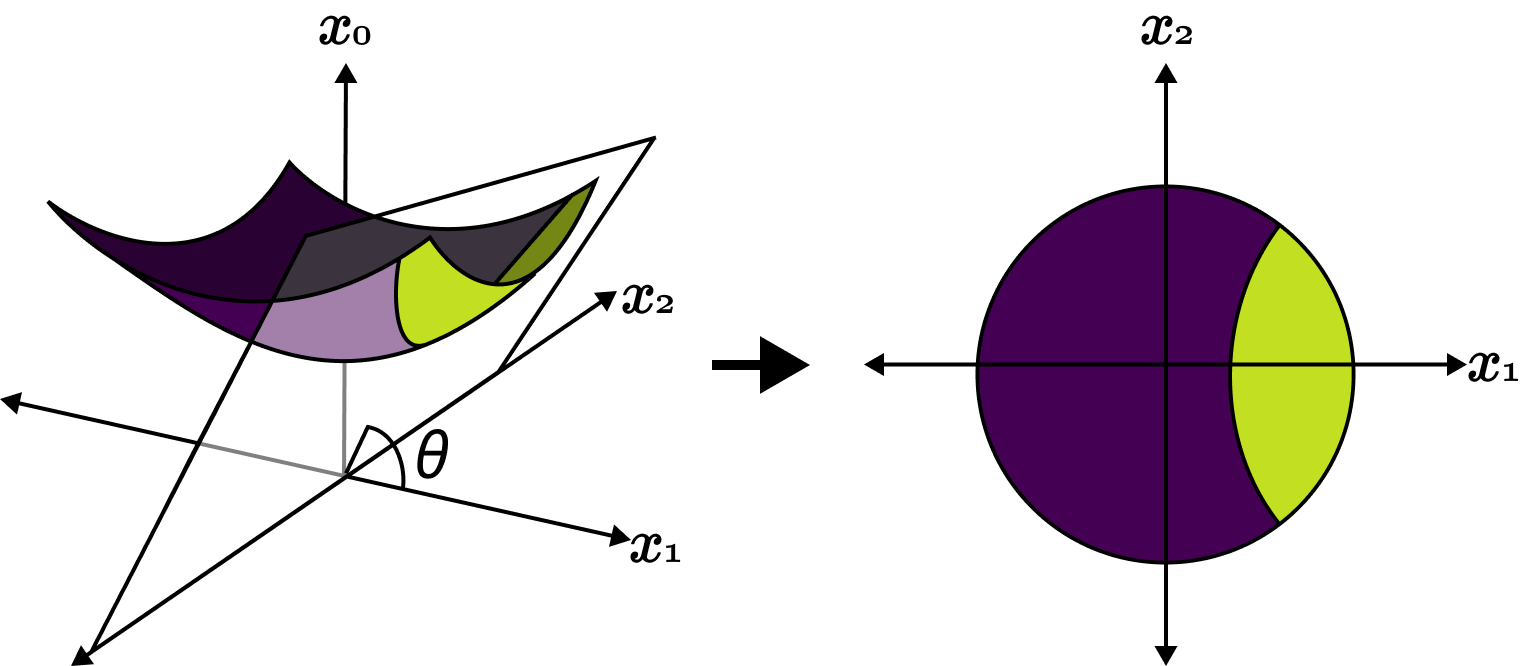}
    \caption{Geodesic partitions in the hyperboloid model $\mathbb{H}^{2,1}$ (left) and poincare model $\mathbb{P}^{2,1}$ (right) into two halves (purple/yellow). In $\mathbb{H}^{2,1}$, a geodesic can be expressed as the intersection of the hyperboloid with an angled plane through the origin of the ambient space (transparent white). While these two representations are equivalent these partitions can be expressed more compactly in $\mathbb{H}^{2,1}$.}
    \label{fig:hyperboloid}
\end{figure}

While hyperbolic classifiers leverage the curvature properties of hyperbolic geometry to make more nuanced and accurate predictions of hierarchical data, such techniques often face a dilemma. Methods employing Riemannian optimization often exhibit sluggishness due to the increased computational complexity associated with operations on Riemannian manifolds, which require intricate geometric calculations. The sensitivity of Riemannian optimization to initialization and the presence of complex geometric constraints can further contribute to slower convergence. Other methods are consistent with hyperbolic geometry but incur time-complexity penalties associated with horosphere calculations (see, for example, ~\citet{fan_horospherical_2023, doorenbos_hyperbolic_2023}). Others apply Euclidean methods to hyperbolic data transformed with logarithmic maps \citep{chami_hyperbolic_2019, chen_fully_2022}, whitening techniques~\citep{chami_horopca_2021}, or directly on hyperbolic coordinates \citep{jiang_learning_2022}. While effective, such methods can ignore the geometric structure of the data or introduce additional complexity to the inference process, adversely affecting both speed and interpretability.

\subsection{The Need for Hyperbolid Decision Trees and Paper Contributions} 

Decision trees are workhorses of machine learning, favored for a variety of reasons, including:
% Decision trees are a fundamental tool in machine learning due to their speed, simplicity and versatility. They possess several advantages that make them a promising candidate for adaptation when designing hyperbolic classifiers. These advantages include: %(i) decision trees are relatively fast to train; (ii) prediction is efficient, making them suitable for real-time or near-real-time applications; and (iii) decision trees are conceptually simple and involve straightforward logic, making them easier to implement and explain than complex neural networks with numerous layers and parameters.
\begin{itemize}
    \item \textit{Speed}. Decision trees are relatively fast to train, and efficient prediction makes them suitable for real-time or near-real-time applications. %Achieving this efficiency necessitates simple decision boundaries that are straightforward to optimize and easy to interpret. 
    \item \textit{Interpretability}. Decision trees provide a highly interpretable model, unlike random forest methods that can be opaque regarding how features combine to form a predictor.
    \item \textit{Simplicity}. Decision trees involve straightforward logic, making them easier to implement and explain than multilayered, highly parameterized network models.
\end{itemize}  
%Despite these advantages, the hypothesis space of decision trees scales exponentially with the number of dimensions, leading to a combinatorial explosion of potential decision boundaries. 
%When burdened with the added complications of hyperbolic geometry, that often requires cumbersome optimization over Riemannian manifolds,there is a gap between the availability of efficient and effective decision tree algorithms for such spaces and their potential impact.
The additional complexities introduced by hyperbolic geometry, which often necessitate intricate optimizations over Riemannian manifolds, introduce a disparity between the existing availability of efficient decision tree algorithms tailored to such spaces and their potential transformative impact.

This paper introduces a novel approach to extend traditional Euclidean decision tree algorithms to hyperbolic space. This approach, called \hyperdt, is conceptually straightforward and maintains constant-time decision complexity while mitigating the scalability issues inherent in high-dimensional Euclidean spaces. Building upon \hyperdt, we introduce an ensemble model called \hyperrf, which is an extension of random forests tailored for hyperbolic space. Our contributions in this work are summarized as follows:
\begin{enumerate}
    \item We develop \hyperdt,\ a novel extension of decision trees to data in hyperbolic space, e.g. learned hyperbolic embeddings. \hyperdt\ avoids computationally expensive Riemannian optimization, numerically unstable exponential or logarithmic maps, and quadratically scaling pairwise comparisons between all data points. Instead, we reframe decision trees in Euclidean space in terms of inner products, yielding a natural extension to hyperbolic spaces of arbitrary negative curvature. Like other decision tree algorithms, \hyperdt\ repeatedly partitions the input space into decision areas, i.e. subspaces labeled with the majority class for the points in the training set for that region. However, because it uses geodesic submanifolds to partition the space, \hyperdt\ is the first decision tree predictor whose decision areas maintain convexity and topological continuity for arbitrary partitions.
    % \item We develop \hyperdt, a novel extension of the decision tree algorithm to data in hyperbolic space, e.g. learned hyperbolic embeddings. \hyperdt\ avoids computationally intensive Riemannian optimization or numerically unstable exponential and logarithmic maps. Instead, we frame Euclidean decision tree algorithms regarding inner products, offering their natural extension to hyperbolic space. This approach works for all negative curvatures and avoids pairwise comparisons between data points. Like other decision tree algorithms, \hyperdt\ repeatedly partitions space into decision areas; however, because it uses geodesic submanifolds to partition the space, \hyperdt\ is the first such predictor whose decision areas are topologically continuous and convex. 
    \item We generalize \hyperdt\ to random forests in a second algorithm which we refer to as \hyperrf. In select cases, \hyperrf\ demonstrates enhanced accuracy and reduced susceptibility to overfitting when contrasted with \hyperdt.
    \item We demonstrate state-of-the-art accuracy and speed of \hyperdt\ and \hyperrf\ on classification problems compared to existing counterparts on various datasets.
    \item We provide a Python implementation of \hyperdt\ and \hyperrf\ for classification and regression following \sklrn\ API conventions \citep{pedregosa_scikit-learn_2011}.
\end{enumerate}

\subsection{Related Work}

Several graph embedding methods have been proposed in the Poincar\'e disk \citep{nickel_poincare_2017, chamberlain_neural_2017}, hyperboloid model \citep{nickel_learning_2018}, and even in mixed-curvature products of manifolds \citep{gu_learning_2019}.
\citet{de_sa_representation_2018} provides a thorough overview and comparison of graph embedding methods in terms of their metric distortions.

Hyperbolic embeddings have found diverse applications across domains, particularly in computational biology and concept ontologies, both of which are structured by latent branching relationships. In biology, inheritance patterns are tree-like: at evolutionary scales, hyperbolic embeddings successfully model well-known phylogenetic trees~\citep{hughes_visualising_2004, chami_trees_2020, jiang_learning_2022}. Furthermore, \citet{corso_neural_2021} learn faithful hyperbolic species embeddings directly from nucleotide sequences, bypassing phylogenetic tree construction. On shorter timescales, such as single-cell RNA sequencing data, where cell types evolve from progenitors, a tree-like structure emerges once more. \citet{ding_deep_2021} showcased that variational autoencoders with a hyperbolic latent space effectively capture the branching patterns of cells' developmental trajectories.

%Hyperbolic embeddings progressed further into the realm of application across several domains, particularly in computational biology and concept ontologies, both of which involve distinct latent structures of data that are tree-like.
%In biological contexts, hyperbolic embeddings are often motivated due to their ability to represent tree-like inheritance patterns. In particular, on an evolutionary scale, they successfully modeled known phylogenetic trees~\citep{hughes_visualising_2004, chami_trees_2020, jiang_learning_2022}.
%Moreover, \citet{corso_neural_2021} showed that faithful hyperbolic species embeddings can be learned directly from nucleotide sequences without first learning a phylogenetic tree.
%On the shorter timescale of single-cell RNA sequencing data, the development of cell types from a progenitor again forms a tree. \citet{ding_deep_2021} show that variational autoencoders with a hyperbolic latent space effectively represent the branching patterns of cells' developmental trajectories.

For concept ontologies like WordNet, hyperbolic embeddings effectively capture subclass relationships among nouns, improving link prediction accuracy \citep{ganea_hyperbolic_2018}. Additionally, \citet{tifrea_poincare_2018} showcase the utility of hyperbolic embeddings for unsupervised learning of latent hierarchies when explicit ontologies are absent, a capability possibly underpinned by the hierarchical nature of concepts within WordNet.
Moreover, a recent study by \cite{desai_hyperbolic_2023} unveils that text-image representations in hyperbolic space exhibit enhanced interpretability and structural organization while maintaining performance excellence in standard multimodal benchmarks.

Machine learning on hyperbolic embeddings is an emerging and evolving research area. 
%Researchers have explored various methods to adapt existing machine learning techniques to hyperbolic spaces. 
\citet{cho_large-margin_2018} and \citep{fan_horospherical_2023} have proposed approaches for adapting support vector machines to hyperbolic space, which has proven useful in biological contexts~\citep{agibetov_using_2019}.
In the realm of neural methods, there have been developments such as hyperbolic attention networks \citep{gulcehre_hyperbolic_2018}, fully hyperbolic neural networks \cite{chen_fully_2022}, and hyperbolic graph convolutional networks \cite{chami_horopca_2021}. A recent contribution by \cite{doorenbos_hyperbolic_2023} introduced $\hororf$, a variant of the random forest method that employs decision trees with horospherical node criteria.
Our approach is most similar to \cite{doorenbos_hyperbolic_2023}; however, it differs in that it utilizes geodesics as opposed to horospheres and does not require pairwise comparisons between data points, leading to an algorithm that maintains constant time decision complexity.%\footnote{ET: Do we want to include organization here? Section 2 provides background on hyperbolic spaces, decision trees, and random forests. Section 3 explains the $\hyperdt$ and $\hyperrf$ algorithms and analyzes its runtime. Section 4 benchmarks these algorithms. Lastly, section 5 contains the conclusion.}

Improving decision tree and random forest algorithms has also attracted considerable research attention, although except for \hororf, none of these methods are designed for use with hyperbolic embeddings. In particular, improvements like gradient boosting~\cite{chen_xgboost_2016} and optimization based on branch-and-bound methods~\cite{lin_generalized_2022, mctavish_fast_2022, mazumder_quant-bnb_2022} have been proposed to circumvent some of the suboptimal qualities of classic CART decision trees~\cite{breiman_classification_2017}, which uses a greedy heuristic to select each split in the tree.

\section{Preliminary}
\subsection{Hyperbolic Spaces}
Hyperbolic geometry, characterized by its constant negative curvature, can be represented by various models, including the hyperboloid (also known as the Lorentz or Minkowski model), the Poincar\'e disk model, the Poincar\'e half-plane model, and the (Beltrami-)Klein model. We use the hyperboloid model due to its simplicity in expressing geodesics using plane geometry, which we exploit to define our decision boundaries. While prior research in this field has predominantly centered on the Poincar\'e disk model, the straightforward conversion between Poincar\'e disk coordinates and hyperboloid coordinates (See Section~\ref{appendix:conversion} for details) allows for seamless integration of techniques across different hyperbolic representations, a flexibility we leverage in our work.

The $D$-dimensional hyperboloid model is embedded inside an ambient $(D+1)$-dimensional Minkowski space, a metric space equipped with the Minkowski inner product:
\begin{equation}
    \langle \x, \x' \rangle_\mathcal{L} = -x_0x_0' + \sum_{i=1}^{D} x_ix_i'\,.
\end{equation}
%In other words, the Minkowski inner product is equivalent to the Euclidean inner product with the first term negated.
%This distinguished dimension is sometimes referred to as the ``timelike'' dimension due to its role in the theory of special relativity.
The above is equivalent to an adaptation of the Euclidean inner product, where the first term is negated. 
This distinguished first dimension, often termed the ``timelike'' dimension, earns its name due to its significance in the context of special relativity theory.

Let the hyperboloid have constant negative scalar curvature $-K$.
Inside the Minkowski space, points are assumed to lie on $\hdk$, the hyperboloid of dimension $D$ and curvature $K$:
\begin{equation}
    \hdk = \left\{ \x \in \mathbb{R}^{D+1} : \langle \x, \x \rangle_\mathcal{L} = -1/K,\ x_0 > 0 \right\}\,.
\end{equation}
That is, the hyperboloid model assumes points lie on the surface of the upper sheet of a two-sheeted hyperboloid embedded in Minkowski space (see Figure \ref{fig:hyperboloid}).
The distance between two points on $\hdk$ is
\begin{equation}
    \delta(\x, \x') = \cosh^{-1}(-K \langle \x, \x'\rangle_\mathcal{L})/\sqrt{K}.
\end{equation}
This distance can be interpreted as the length of the geodesic, the shortest path on the manifold connecting $x$ and $x'$.
%This path is also known as the geodesic.
In the hyperboloid model, all geodesics are intersections of $\hdk$ with respective 2D planes that pass through the origin of the Minkowski space.

\subsection{Decision Tree Algorithms}
% \paragraph{Decision Tree Algorithms.} 
%\paragraph{Classification and Regression Trees.} 
The {\em Classification and Regression Trees} (CART) algorithm is a mainstay of machine learning, alongside extensions such as random forests \citep{breiman_random_2001} and XGBoost \citep{chen_xgboost_2016}. 
CART recursively partitions the feature space into increasingly homogenous subspaces by maximizing the information gain at each split $S$, $IG(S)$. 
It measures improved homogeneity due to splitting a dataset $\mathbf{X}$ into subsets $\mathbf{X}^0,\mathbf{X}^1$ of respective fractions $f^i = |\mathbf{X}^i|/\mathbf{X}$:
\begin{equation}
IG(S) = C(\mathbf{X}) - f^0 C(\mathbf{X}^0) - f^1 C(\mathbf{X}^1),
\end{equation}
where $C(\cdot)$ is the impurity or cost function of each set.
Objective functions like Gini impurity, mean squared error (MSE), and entropy are popular choices for $C(\cdot)$.
% \begin{align}
% \text{Gini:} \quad G(D) &= 1 - \sum_{j=1}^{m} p_j^2\\
% \text{MSE:} \quad E(D) &= \frac{1}{|D|} \sum_{i=1}^{|D|} (y_i - \bar{y})^2
% \end{align}
%After creating the tree, it can be pruned to avoid overfitting. 
%The final tree can be used for making predictions on new data points.
%It is common to train random forests, which are ensembles of decision trees on randomly subsampled versions of the training data.
%In this case, each component tree is trained on a slightly different view of the training data, and their predictions can be aggregated together to yield a single prediction.
The tree is iteratively constructed until further splits will constitute overfitting. %The resulting tree is then ready for making predictions on new data points.
A decision tree is often used as is to make predictions. 
The {\em Random Forest} algorithm integrates predictions across ensembles of decision trees trained on randomly subsampled subsets of the training data using a majority voting procedure.

\section{Hyperboloid Decision Tree Algorithms}
Extending CART to hyperbolic space involves several essential steps. First, as outlined in Section~\ref{innerproduct}, we express Euclidean decision boundaries in terms of inner products, providing the requisite geometric intuition for hyperbolic decision trees. In Section~\ref{hyperboloidmodel}, we utilize these inner products to establish a streamlined decision process based on geodesic submanifolds in hyperbolic space. Section~\ref{choosinghyperplanes} discusses the selection of candidate hyperplanes, Section ~\ref{boundaries} presents closed-form equations for decision boundaries, and Section ~\ref{randomforest} describes the Hyperbolic Random Forest extension \hyperrf.
%Finally, Section~\ref{furtherdevelopments} presents extensions and closed-form equations for decision boundaries. 

\subsection{Formulating Decision Trees with Inner Products}
\label{innerproduct}
%Before we define our model, we need to make one additional observation about the CART algorithm.
%A split is typically thought of as determining whether the value of a point $x$ at some dimension $d$ is greater or less than a threshold $\theta$, i.e.
%Before delving into the model's definition, it is essential to make an additional observation regarding the CART algorithm. 
Traditionally, a split is perceived as a means to ascertain whether the value of a point $\x$ in a given dimension $d$, is greater or lesser than a designated threshold, $\theta$, namely,
\begin{equation}
\label{eq:threshold_decision}
    S(\x) = \mathbb{I}\{ x_d > \theta\}
    \,.
\end{equation}
This decision boundary can also be thought of as the axis-parallel hyperplane $x_d = \theta$, and thus we can rewrite the same split as follows
\begin{equation}
\label{eq:dot_product_decision}
    S(x) = \max(0,\ \text{sign}(\x \cdot \nd - \theta))\,,
\end{equation}
where $\nd$ is the one-hot base vector along dimension $d$, i.e. the 
normal vector of our decision hyperplane. 
Of course, Eq.~\ref{eq:threshold_decision} is the practical, $O(1)$ condition. The slower, $O(D)$ Eq.~\ref{eq:dot_product_decision} is instructive towards the hyperboloid generalization, and can still be computed in $O(1)$ due to sparsity of $\nd$.

%Computing inner products is impractical because it requires $O(D)$ time, whereas checking a threshold value is $O(1)$.
%However, sparsity allows $x \cdot \vec{n_d}$ to be expressed simply as as $x_d$, and therefore $S = \text{sign}(x_d - \theta)$, which can be computed in $O(1)$ time.

\subsection{Extension to the Hyperboloid Model}
\label{hyperboloidmodel}

We will now modify decision tree splits for hyperbolic space. Splitting a decision tree along standard axis-aligned hyperplanes is inappropriate when all the datapoints lie on the hyperboloid: the intersection of an axis-aligned $D+1$ dimensional hyperplane with the hyperboloid $\hdk$ in $(D+1)$-dimensional Minkowski space results in a $D$-dimensional hyperbola, which lacks any meaningful interpretation within the hyperboloid model. Euclidean CART generates such decision boundaries, but they are likely ill-suited to capture the geometry of hyperbolic space.

%{\textcolor{red}{[Eitan:] Maybe include a diagram similar to Figure 1 left which shows an axis aligned hyperplane intersecting the hyperboloid in a non-sensical way. Specifically, having a hyperplane parallel to $x_0$ would be partciularly illustrastive of a split that does not make sense in hyperbolic space.}}

% We will now introduce customized decision tree splits that suits the geometry of the hyperboloid model better than the standard axis-aligned hyperplanes.
% This is needed because such a standard hyperplane 
% in $(D+1)$-dimensional Minkowski space,
% intersects $\mathbb{H}^D$ in $(D-1)$-dimensional manifold which is a $D$-dimensional hyperbola, which does not have a special interpretation within the hyperboloid model. Consequently, decision boundaries generated by Euclidean CART within the hyperbolic space are likely ill-suited to capture its geometry.

On the other hand, $D$-dimensional homogeneous hyperplanes, i.e. hyperplanes that contain the origin, intersect $\hdk$ as geodesic submanifolds.
In 3D Minkowski space, geodesics between any $\{\x, \x'\} \subset \mathbb{H}^{2,1}$ lie on the intersection of $\mathbb{H}^{2,1}$ with some homogeneous 2D plane, as in  Figure~\ref{fig:hyperboloid}.
 %In 3D, for any two points, $x$ and $x'$, located on the intersection of a hyperboloid with such a plane, the geodesic path between $x$ and $x'$ lies entirely within that intersection. 
 In higher dimensions, homogeneous hyperplanes intersect $\hdk$ as $(D-1)$-dimensional geodesic submanifolds, which likewise contain all geodesics between their elements. Partitions by homogeneous hyperplanes maintain convexity and topological continuity: all pairs of points in a subspace are reachable by shortest paths that stay completely within their own subspace.  
 
Building upon the inner product formulation of splits Euclidean CART (Eq.~\ref{eq:dot_product_decision}),
we can substitute a set of geometrically appropriate decision boundaries without altering the rest of the CART framework.
%upon the observation that thresholding in Euclidean CART can be represented by the sign of the inner product of a point with the normal vector of an axis-aligned hyperplane, We can substitute a set of decision boundaries that align with the geometry. 
Specifically, we replace axis-parallel hyperplanes with homogeneous ones. 

To maintain the dimension-by-dimension character of Euclidean CART and enforce sparse normal vectors for efficient inner product calculations, we further restrict the number of decision boundary candidates to $O(D|\mathbf{X}|)$ by only considering rotations of the plane $x_0 = 0$ along a single other axis $d$. These hyperplanes are fully parameterized by $d$ and the rotation angle $\theta$, yielding corresponding normal vectors
\begin{equation}
    \label{eq:normal}
    \ndt := \langle n_0 = -\cos(\theta), 0, \ldots, 0, n_d=\sin(\theta), 0, \ldots, 0 \rangle.
\end{equation} 
They define hyperplanes that satisfy 
\begin{equation}
    x_0 \cos(\theta) - x_d\sin(\theta) = 0
\end{equation}
The sparsity of $\ndt$ yields a compact $O(1)$ decision procedure:
\begin{equation}
    \label{sparse_dp_hyperboloid}
    S(x) = \text{sign}\left(\max\left(0,\ \sin(\theta)x_d - \cos(\theta)x_0\right)\right)
\end{equation}
Notably, this procedure determines points' position relative to a geodesic decision boundary without computing the actual location of the geodesic on $\hdk$. Because of this, it is also curvature-agnostic. Hyperbolic decision trees compose splits analogously to Euclidean CART: the same objective functions are applicable, and so is the consideration of a single candidate decision boundary per point per (space-like) dimension, resulting in identical asymptotic complexity.
%It's worth noting that since the timelike dimension $x_0$ is included in each boundary evaluation, only the $D$ spacelike dimensions are taken into account as potential decision boundary candidates.

%The rest of the model is analogous to Euclidean CART.
%The same objective functions apply, and one decision boundary per point per dimension is typically considered, resulting in no change in asymptotic complexity.
%Since the timelike dimension $x_0$ is included in each boundary evaluation, only the $d$ spacelike dimensions are considered as decision boundary candidates.

%Note that the preceding discussion used Euclidean dot-products even though Minkowski space uses the Minkowski dot-product in reality.
%We expect that Euclidean dot products are more familiar and intuitive for most readers and therefore are more helpful for understanding this model.
%In the Appendix\footnote{PC: need to make a ref here} we demonstrate the equivalence between models based on Minkowski and Euclidean inner products. 
%% ANTONIO: I've replaced all references of the dot product with inner product, as this is more generalizable and is defined for Minkowski space.

\subsection{Choosing Candidate Hyperplanes}
\label{choosinghyperplanes}
In Euclidean CART, candidate thresholds are classically chosen among midpoints between successive observed $x_d$ values in the data.
The hyperbolic case is slightly more nuanced.

Each decision boundary for dimension $d$ is parameterized by an angle $\theta$, instead of a coordinate value. Each point $\mathbf{x}$ lies on a plane of angle $\theta = \tan^{-1}(x_0 / x_d)$. The midpoint angle $\theta_m$ between two angles $\theta_1 < \theta_2$ is defined in terms of points lying on the intersection of these hyperplanes with $\hdk$.
% In particular, we want points on the plane corresponding to $\theta_m$ to be equidistant to corresponding points on planes corresponding to $\theta_1$ and $\theta_2$. 
By setting all dimensions besides 0 and $d$ to zero, we can solve for the angle corresponding to the point on $\hdk$ that is exactly equidistant to the points corresponding to angles $\theta_1, \theta_2$:
\begin{equation}
    \theta_m = \begin{cases} 
                \cot^{-1}\left( V -\sqrt{V^2 - 1} \right) &\text{ if } \theta_1 < \pi - \theta_2\\
        \cot^{-1}\left( V +\sqrt{V^2 - 1} \right) &\text{ if } \theta_1 > \pi - \theta_2
    \end{cases} \label{eq:midpoint}
\end{equation}
where $ V := \frac{\sin(2\theta_21- 2\theta_2)}{2\sin(\theta_1 + \theta_2)\sin(\theta_2 - \theta_1)}$.
See Appendix Section~\ref{appendix:midpoints}  for a full derivation.

% \subsection{Further Developments and Methods}
% \label{furtherdevelopments}
% \paragraph{Hyperboloid Random Forest.}

% \paragraph{Parameterizing Decision Boundaries.}
\subsection{Parameterizing Decision Boundaries}
\label{boundaries}
Let $\mathbf{P}(\theta)$ be a decision hyperplane learned by \hyperdt\ (without loss of generality, assume $d=1$). We derive closed-form equations for the geodesic submanifold where $\mathbf{P}(\theta)$ intersects $\hdk$.
When all other dimensions are $0$, this occurs when
\begin{equation}
    x_0 = \alpha(\theta, K)\sin(\theta);\ 
    x_1 = \alpha(\theta, K)\cos(\theta),
\end{equation}
where $\alpha(\theta, K) := \sqrt{-\sec(2\theta)} / \sqrt{K}$.
Note that this is also the point on the intersection of $\mathbf{P}(\theta)$ and $\hdk$ that is closest to the origin. 
We use this to parameterize the entire geodesic submanifold $\mathbf{G^d}$ resulting from intersecting the plane with $\hdk$:
% We use this to parameterize an entire geodesic arc $\mathbf{g}(t)$
% along any additional space-like dimension $d'\neq d$:
% \begin{equation}
%     g_0(t) = \cosh(t)\ \alpha(\theta, K)\sin(\theta);\ 
%     g_d(t) = \cosh(t)\ \alpha(\theta, K)\cos(\theta);\
%     g_{d'}(t) = \sinh(t) / \sqrt{K}
% \end{equation}
% Combining such multiple such additional dimensions allows parameterization of the entire geodesic submanifold.
\begin{align}
    \mathbf{v^0} &= \langle \sin(\theta),\ \cos(\theta),\ 0,\ \ldots \rangle\\
    \mathbf{u^d} &= \langle 0,\ \ldots,\ u_d^d=1,\  \ldots,\ 0 \rangle,\ 2 \leq d \leq D\\
    % \mathbf{g^1}(\theta, K,  t) &= \cosh(t)\cdot\alpha(\theta, K)\cdot\mathbf{v^0} + \sinh(t)\mathbf{u^2}/\sqrt{K}\\
    % \mathbf{G^1}(\theta, K) &= \left\{\mathbf{g^1}(\theta, K, t)\ :\ t \in \mathbb{R} \right\}\\
    \mathbf{G^1}(\theta, K) &= \left\{ \cosh(t)\cdot\alpha(\theta, K) \cdot \mathbf{v^0} + \sinh(t)\mathbf{u^2} / \sqrt{K} : t \in \mathbb{R} \right\}\\
    \mathbf{G^{d}}(\theta, K) &= \left\{\cosh(t)\mathbf{v^{d-1}} + \sinh(t)\mathbf{u^{d+1}} / \sqrt{K} : \mathbf{v^{d-1}} \in \mathbf{G^{d-1}}(\theta, K),\ t \in \mathbb{R} \right\}
\end{align}
For visualization, we use $\mathbb{H}^{2,1}$ projected to the Poincar\'e disk $\mathbb{P}^{2,1}$.
We recursively partition the space, plotting decision boundaries at each level and coloring the partitioned space by the majority class.
Figure~\ref{fig:visualization} shows an example of such a plot. See Appendix Section~\ref{appendix:visualization} for a full derivation.

\begin{figure}
    \centering
    \includegraphics[width=\textwidth]{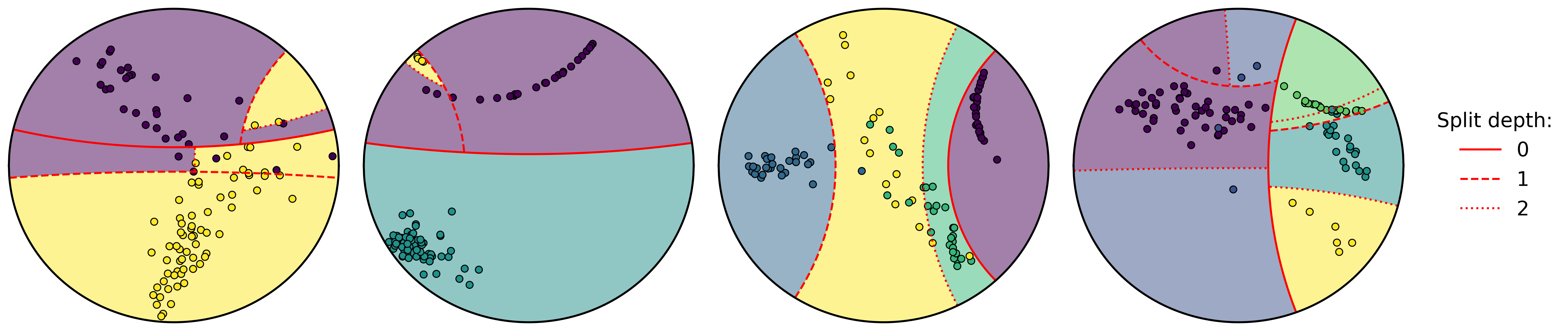}
    \caption{ %Decision boundaries visualized on the Poincar\'e disk for real hyperboloid decision trees in 2, 3, 4, and 5-class classification settings. The data points are sampled from a mixture of wrapped normal distributions, and all trees have a maximum depth of 3. The decision tree does not undergo pruning after training. The color of the region is the predicted class; the color of the datapoint is the true class.
    Learned \hyperdt\ decision boundaries for 2, 3, 4, and 5-class mixtures of wrapped normal distributions visualized on the Poincar\'e disk. All trees have a maximum depth of 3 and forgo post-training pruning. In the visualization, regions are colored according to their \textit{predicted} class labels while data points are colored according to their \textit{true} class labels.}
    \label{fig:visualization}
\end{figure}

\subsection{Hyperboloid Random Forests}
\label{randomforest}
Analogously to the Euclidean case, creating a hyperboloid random forest is possible by training an ensemble of hyperboloid decision trees on randomly resampled versions of the training data.
For speed, we implement \hyperrf, a multithreaded version of hyperboloid random forests wherein each tree is trained as a separate process.

\section{Classification Experiments}%S\footnote{A separate set of experiments and results, for a regression extension of our framework, is reported only in Section~\ref{appendix:image_embeddings}, due to lack of space.}

\subsection{Performance Benchmark
%}
%\subsubsection{
Baselines}
For decision trees, we compare our method to standard Euclidean decision trees as implemented in \sklrn~\citep{pedregosa_scikit-learn_2011}.
Since we use the same parameters as the \sklrn\ decision tree and random forest classes, we can ensure that implementation details like maximum depth and number of points in a leaf node can be standardized.
% The same extends to random forests.
We additionally compare our random forest method to $\hororf$\ \citep{doorenbos_hyperbolic_2023}, another ensemble classifier for hyperbolic data.

Since \hororf\ does not implement a single decision tree method, we modified its code to avoid resampling the training data when training a single tree. We call this version \horodt.

For all predictors, we use trees with depth $\leq 3$ and $\geq 1$ sample per leaf.
For random forests, all methods use an ensemble of 12 trees.
We explore performance in $D=2$, 4, 8, and 16 dimensions.

%\subsubsection{
\subsection{
Datasets}

\paragraph{Synthetic Datasets.}
We create a hyperbolic mixture of Gaussians, the canonical synthetic dataset for classification benchmarks, following \citet{cho_large-margin_2018}.
We use the wrapped normal distribution on the hyperboloid for each Gaussian as described in \cite{nagano_wrapped_2019}.
We draw the means of each Gaussian component from a normal distribution in the tangent plane at the origin and project it onto the hyperboloid directly using an exponential map.
The timelike components of random Gaussians may grow quite large in high dimensions, creating both numerical instability and trivially separable clusters. We thus shrink the covariance matrix $D$-fold. See Section~\ref{appendix:gaussians} for details.
% For $\hororf$ benchmarks, we project these samples from $\mathbb{H}^D$ to a $D$-dimensional Poincar\'e disk.

\paragraph{NeuroSEED RNA Embeddings.}
NeuroSEED \citep{corso_neural_2021} is a method for embedding DNA sequences into a (potentially hyperbolic) latent space by using a Siamese neural network with a distance-preserving loss function.
This method encodes DNA sequences directly into latent space without first constructing a phylogenetic tree---or, equivalently, it embeds a dense graph of pairwise edit distances.
We trained 2, 4, 8, and 16-dimensional Poincar\'e embeddings of the 1,262,987 16S ribosomal RNA sequences from the Greengenes database \citep{mcdonald_greengenes2_2023}, then filtered them to the 37,215 that have been identified in the American Gut Project \cite{mcdonald_american_2018}.
This downsampling yields
%restricts Greengenes to species that have been identified in the human gut, resulting in 
a clinically relevant subset of all Prokaryote species.
%This dataset has rich taxonomic annotations, frequently down to the species level.
For the purposes of these benchmarks, however, 
we restricted ourselves to predicting the six most abundant phyla: Firmicutes, Proteobacteria, Bacteroidetes, Actinobacteria, Acidobacteria, and Plantomycetes.
% For \sklrn\ and \horodt\ methods, we project the learned $D$-dimensional Poincar\'e embeddings to $\mathbb{H}^D$.

\paragraph{Polblogs Graph Embeddings.}
We use Polblogs \citep{adamic_political_2005}, a canonical dataset in the hyperbolic embeddings literature for graph embeddings.
In the Polblogs dataset, nodes represent political blogs during the 2004 United States presidential election, and edges represent hyperlinks between blogs.
Each blog is labeled according to its political affiliation, ``liberal'' or ``conservative.''
We use the \texttt{hypll}~\citep{van_spengler_hypll_2023} Python implementation of the \citet{nickel_poincare_2017} method to compute 10 randomly initialized Poincar\'e disk embeddings in 2, 4, 8, and 16 dimensions.
% For \sklrn\ and \horodt\ methods, we project the learned $D$-dimensional Poincar\'e embeddings to $\mathbb{H}^D$.

%\subsubsection{
\subsection{
Benchmarking Procedure}

We benchmarked our method against Euclidean random forests as implemented in \sklrn, and against $\hororf$.
Each predictor was run with the same settings, including dataset and random seed.
$\hororf$ and $\horodt$ used Poincar\'e disk coordinates and all other modelsused hyperboloid model coordinates.
Each dataset was converted to the appropriate model of hyperbolic space for its predictor before training.

For Gaussian and NeuroSEED datasets, we drew 100, 200, 400, and 800 samples using the same five seeds.
We recorded micro- and macro-averaged F1 scores, AUPRs, and run times under 5-fold cross-validation.
Cross-validation instances were seeded identically across predictors.
We could not produce per-fold timing for $\hororf$, so instead we recorded the time it took to run the full 5-fold cross-validation.
For fairness, we timed the data loading scripts separately and subtracted these times from the reported $\hororf$ times.

We conducted paired, two-tailed t-tests comparing each pair of predictors and marked 
significant differences in Table \ref{tab:accuracies}. A full table of $p$-values is available in Section \ref{appendix:p_vals} in the Appendix.

Benchmarks were conducted on an Ubuntu 22.04 machine equipped with an Intel Core i7-8700 CPU (6 cores, 3.20 GHz), an NVIDIA GeForce GTX 1080 GPU with 11 GiB of VRAM, and 15 GiB of RAM.
Storage was handled by a 2TB HDD and a 219GB SSD.
Experiments were implemented using Python 3.11.4, accelerated by CUDA 11.4 with driver version 470.199.02.

\subsection{Results}
%\subsection{
\begin{table}[t]
    \begin{center}
    %\scalebox{0.7}{
    \begin{tabular}{lll|lll|lll}
\toprule
 &  &  & \multicolumn{3}{c|}{Decision Trees} & \multicolumn{3}{c}{Random Forests} \\
Data & $D$ & $n$ & \hyperdt & \textsc{Sklearn} & \horodt & \hyperrf & \textsc{Sklearn} & \hororf \\
%Data & $d$ & $n$ &  &  &  &  &  &  \\
\midrule
\multirow[t]{16}{*}{
\begin{sideways}
\hspace{-3.5cm}
Gaussian
\end{sideways}
} & \multirow[t]{4}{*}{2} & 100 & \textbf{89.10}\textsuperscript{$\dagger$} & 87.90 & 84.60 & \textbf{90.70}\textsuperscript{$\ddagger$}\textsuperscript{$\dagger$} & 87.50 & 86.30 \\
 &  & 200 & \textbf{90.05}\textsuperscript{$\dagger$} & 89.55 & 84.60 & \textbf{90.60} & 89.15 & 89.10 \\
 &  & 400 & \textbf{90.97}\textsuperscript{$\ddagger$}\textsuperscript{$\dagger$} & 89.53 & 85.55 & \textbf{91.32}\textsuperscript{$\ddagger$}\textsuperscript{$\dagger$} & 89.00 & 88.88 \\
 &  & 800 & \textbf{91.88}\textsuperscript{$\ddagger$}\textsuperscript{$\dagger$} & 90.14 & 85.75 & \textbf{91.99}\textsuperscript{$\ddagger$}\textsuperscript{$\dagger$} & 89.33 & 89.45 \\
\cline{2-9}
 & \multirow[t]{4}{*}{4} & 100 & \textbf{98.70}\textsuperscript{$\dagger$} & 97.70 & 93.60 & \textbf{98.40} & 97.90 & 97.90 \\
 &  & 200 & \textbf{98.75}\textsuperscript{$\ddagger$}\textsuperscript{$\dagger$} & 98.10 & 95.80 & \textbf{98.85}\textsuperscript{$\ddagger$}\textsuperscript{$\dagger$} & 97.90 & 98.05 \\
 &  & 400 & \textbf{99.25}\textsuperscript{$\ddagger$}\textsuperscript{$\dagger$} & 98.25 & 96.92 & \textbf{99.30}\textsuperscript{$\ddagger$}\textsuperscript{$\dagger$} & 98.22 & 98.50 \\
 &  & 800 & \textbf{99.30}\textsuperscript{$\ddagger$}\textsuperscript{$\dagger$} & 98.36 & 97.27 & \textbf{99.36}\textsuperscript{$\ddagger$}\textsuperscript{$\dagger$} & 98.21 & 98.76 \\
\cline{2-9}
 & \multirow[t]{4}{*}{8} & 100 & \textbf{99.70}\textsuperscript{$\dagger$} & 99.60 & 97.70 & \textbf{99.70} & 99.50 & 99.10 \\
 &  & 200 & \textbf{99.65}\textsuperscript{$\dagger$} & 99.60 & 98.20 & \textbf{99.75} & 99.70 & \textbf{99.75} \\
 &  & 400 & \textbf{99.90}\textsuperscript{$\dagger$} & 99.88 & 99.10 & 99.88 & \textbf{99.93} & 99.88 \\
 &  & 800 & \textbf{99.96}\textsuperscript{$\dagger$} & 99.90 & 99.38 & \textbf{99.96} & 99.91 & 99.94 \\
\cline{2-9}
 & \multirow[t]{4}{*}{16} & 100 & \textbf{99.80}\textsuperscript{$\dagger$} & 99.50 & 98.80 & \textbf{99.80} & 99.60 & 99.60 \\
 &  & 200 & 99.95 & \textbf{100.00}\textsuperscript{$\dagger$} & 99.50 & 99.90 & \textbf{99.95} & 99.80 \\
 &  & 400 & \textbf{100.00}\textsuperscript{$\dagger$} & 99.97 & 99.90 & \textbf{100.00} & \textbf{100.00} & 99.95 \\
 &  & 800 & \textbf{100.00} & 99.99 & 99.90 & \textbf{100.00} & 99.99 & 99.92 \\
\cline{1-9} \cline{2-9}
\multirow[t]{16}{*}{
\begin{sideways}
\hspace{-3.5cm}
NeuroSEED
\end{sideways}
} & \multirow[t]{4}{*}{2} & 100 & \textbf{56.60}\textsuperscript{$\dagger$} & 55.60 & 49.70 & \textbf{57.20} & 55.70 & 56.80 \\
 &  & 200 & \textbf{59.60}\textsuperscript{$\ddagger$}\textsuperscript{$\dagger$} & 58.45 & 50.35 & 60.10 & 58.20 & \textbf{60.25} \\
 &  & 400 & \textbf{61.78}\textsuperscript{$\ddagger$}\textsuperscript{$\dagger$} & 61.00 & 50.62 & \textbf{61.58}\textsuperscript{$\ddagger$}\textsuperscript{$\dagger$} & 59.47 & 59.33 \\
 &  & 800 & \textbf{61.69}\textsuperscript{$\dagger$} & 61.68 & 54.11 & \textbf{62.05}\textsuperscript{$\ddagger$}\textsuperscript{$\dagger$} & 59.75 & 59.94 \\
\cline{2-9}
 & \multirow[t]{4}{*}{4} & 100 & \textbf{80.40}\textsuperscript{$\dagger$} & 80.30 & 53.40 & \textbf{80.90}\textsuperscript{$\dagger$} & 79.20 & 71.50 \\
 &  & 200 & 83.60 & \textbf{83.70}\textsuperscript{$\dagger$} & 52.70 & \textbf{84.45}\textsuperscript{$\ddagger$}\textsuperscript{$\dagger$} & 82.00 & 70.40 \\
 &  & 400 & \textbf{83.88}\textsuperscript{$\dagger$} & 83.83 & 54.08 & \textbf{84.65}\textsuperscript{$\ddagger$}\textsuperscript{$\dagger$} & 82.33 & 65.20 \\
 &  & 800 & 84.49 & \textbf{84.50}\textsuperscript{$\dagger$} & 55.69 & \textbf{84.96}\textsuperscript{$\ddagger$}\textsuperscript{$\dagger$} & 82.03 & 65.70 \\
\cline{2-9}
 & \multirow[t]{4}{*}{8} & 100 & 73.80 & \textbf{74.00}\textsuperscript{$\dagger$} & 52.60 & 79.50 & \textbf{82.80}\textsuperscript{*}\textsuperscript{$\dagger$} & 70.80 \\
 &  & 200 & 78.30 & \textbf{78.40}\textsuperscript{$\dagger$} & 55.35 & 81.40 & \textbf{84.35}\textsuperscript{*}\textsuperscript{$\dagger$} & 65.55 \\
 &  & 400 & 79.45 & \textbf{79.57}\textsuperscript{$\dagger$} & 52.30 & 82.42 & \textbf{86.40}\textsuperscript{*}\textsuperscript{$\dagger$} & 62.88 \\
 &  & 800 & \textbf{80.76}\textsuperscript{$\dagger$} & \textbf{80.76}\textsuperscript{$\dagger$} & 50.55 & 82.03 & \textbf{86.34}\textsuperscript{*}\textsuperscript{$\dagger$} & 57.69 \\
\cline{2-9}
 & \multirow[t]{4}{*}{16} & 100 & \textbf{74.10}\textsuperscript{$\dagger$} & 73.50 & 61.80 & 80.70 & 79.90 & \textbf{83.30} \\
 &  & 200 & \textbf{75.90}\textsuperscript{$\dagger$} & 75.55 & 66.75 & 82.70 & 82.55 & \textbf{83.85} \\
 &  & 400 & \textbf{77.05}\textsuperscript{$\dagger$} & 77.03 & 68.80 & 82.30 & 84.73 & \textbf{85.90}\textsuperscript{*} \\
 &  & 800 & 79.21 & \textbf{79.22}\textsuperscript{$\dagger$} & 69.59 & 82.44 & 84.44 & \textbf{85.49}\textsuperscript{*} \\
\cline{1-9} \cline{2-9}
\multirow[t]{4}{*}{
\begin{sideways}
\hspace{-1.2cm}
Polblogs
\end{sideways}
} & 2 & 979 & \textbf{71.04}\textsuperscript{$\dagger$} & 70.35 & 64.73 & 71.40 & \textbf{71.65}\textsuperscript{$\dagger$} & 66.33 \\
% \cline{2-9}
 & 4 & 979 & \textbf{71.53}\textsuperscript{$\dagger$} & 70.83 & 62.38 & \textbf{72.31}\textsuperscript{$\dagger$} & 72.10 & 68.53 \\
% \cline{2-9}
 & 8 & 979 & 74.02 & \textbf{74.85}\textsuperscript{*}\textsuperscript{$\dagger$} & 61.58 & 74.87 & \textbf{75.36}\textsuperscript{$\dagger$} & 63.60 \\
% \cline{2-9}
 & 16 & 979 & \textbf{75.06}\textsuperscript{$\dagger$} & \textbf{75.05}\textsuperscript{$\dagger$} & 63.70 & 76.36 & \textbf{76.80}\textsuperscript{$\dagger$} & 69.12 \\
% \cline{1-9} \cline{2-9}
\bottomrule
\end{tabular}
   
    %}
    \end{center}
    \caption{Mean micro-F1 scores for classification benchmarks over 10 seeds and 5 folds. The highest-scoring decision tree and random forests are bolded separately. $*$ means a predictor beat \hyperrf, $\dagger$ means a predictor beat $\hororf$, and $\ddagger$ means a predictor beat \sklrn, with $p < 0.05$.}
    \label{tab:accuracies}
\end{table}
\paragraph{Classification Scores.}
% \footnote{Can we include STDs in this table? Would be nice to color code this table in order of increasing values. Check this table from NeruoSeed out as an example of what we want: \url{https://github.com/gcorso/NeuroSEED/blob/master/tutorial/hc_average.png} }
% Out of 36 distinct dataset, dimension, and sample size combinations, our method was the best 26 times, of which 3 were 2-way ties with scikit-learn and 2 were 3-way ties.
% Furthermore, we demonstrated statistically significant advantages over $\hororf$ in 9 cases and scikit-learn in 8 cases. 

% The next-best predictor was scikit-learn, which was the best 10 times, 3 of which were 2-way ties and 2 of which were 3-way ties.
% Of these, they achieved statistically significant advantages over $\hororf$ in 5 cases, and over our method in 3 cases (all of which were on the NeuroSEED 8-dimensional dataset).

% Finally, $\hororf$ won in 7 cases, of which 2 were 3-way ties. 
% $\hororf$ statistically outperformed our method twice on the 16-dimensional NeuroSEED embeddings.
The results of the classification benchmark are summarized in Table \ref{tab:accuracies}.
Out of 36 distinct dataset, dimension, and sample size combinations, \hyperdt\ had the highest score 28 times (one of which was a tie with $\sklrn$). We demonstrated a statistically significant advantage over $\sklrn$ decision trees in 7 cases, and over HoroRF in 27 cases. \sklrn\ statistically outperformed \hyperdt\ once, on the 8-dimensional Polblogs dataset.

Similarly, \hyperrf\ won 22 times, tying once each with \sklrn\ and HoroRF, and statistically outperforming \sklrn\ in 11 cases and $\hororf$ in 13 cases. \sklrn\ statistically outperformed \hyperrf\ in 4 cases, all on 8-dimensional NeuroSEED data, and HoroRF statistically outperformed \hyperrf\ in 2 cases, both on 16-dimensional NeuroSEED data.

Overall, both \hyperdt\ and \hyperrf\ showed substantial advantages over comparable methods on the datasets and hyperparameters tested. 
%This suggests that geodesic decision boundaries are well-suited to certain types of hyperbolic data, and can provide a classification advantage. 
In high dimensions, classifiers tended to converge to uniformly high performance. The best model for NeuroSEED embeddings varied by dimensionality, a phenomenon that warrants further investigation.

\begin{figure}%[h]
    \begin{center}
        \includegraphics[width=\textwidth]{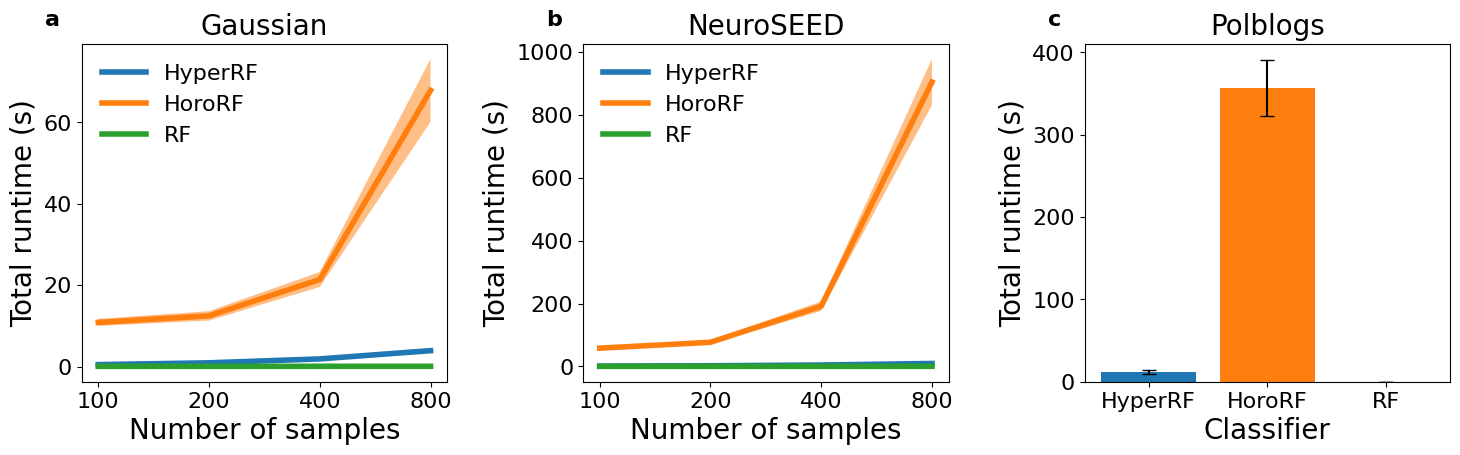}
    \end{center}
    \caption{Time to run 5-fold cross-validation, averaged over 10 seeds for each classifier as a function of the number of points. Shaded regions are 95\% confidence intervals. Split by dataset: \textbf{(a)} wrapped normal mixture, \textbf{(b)} NeuroSEED OTU embeddings, and \textbf{(c)} Polblogs embeddings.}
    \label{fig:runtimes}
\end{figure}

%\subsection{
\paragraph{Runtime Analysis.}
In addition to accuracies, we report runtimes for each classifier on each task.
In particular, we are interested in the asymptotic behavior of our predictors as a function of the number of samples being considered.
These runtimes are plotted by dataset in Figure \ref{fig:runtimes}.
We demonstrate that our method, while slower than the $\sklrn$ implementation by a constant factor, is always faster than $\hororf$ and grows linearly in runtime with the number of samples.
In contrast, $\hororf$ grows quadratically in runtime as the number of samples increases.
This is likely due to the additional complexity of learning horospherical decision boundaries using HoroSVM.

Because $\hororf$ is optimized for GPU, whereas $\hyperrf$ is optimized for parallelism on CPU, exact runtime ratios can be misleading and highly machine-dependent. Similarly, $\hyperrf$ may lack some optimizations found in $\sklrn$. Therefore, we emphasize the asymptotic aspect of this benchmark, which is agnostic to hardware details and constant-time optimizations. 

\paragraph{Additional experiments}
The results of additional experiments can be found in the Appendix. Section \ref{appendix:scaling} contains additional scaling benchmarks. Sections \ref{appendix:hsvm_hmlr} and \ref{appendix:other_geometries} extend benchmarks to other hyperbolic classifiers and other models of hyperbolic geometry, respectively. Sections \ref{appendix:image_embeddings} and \ref{appendix:wordnet} test performance on image and text embeddings, respectively. Finally, Section \ref{appendix:ablation} tests the impact of ablating the midpoint computation described in Equation \ref{eq:midpoint}.

\section{Conclusion}
We have introduced $\hyperdt$, a novel formulation of decision tree algorithms tailored for hyperbolic spaces. This approach leverages inner products to establish a streamlined decision procedure via geodesic submanifolds. \hyperdt\ exhibits constant-time evaluation at each split and does not rely on Riemannian optimization nor pairwise point comparisons in training or prediction. We extended this technique to random forests with \hyperrf, providing versatile tools for classification and regression tasks. \hyperdt\ is more accurate than analogous methods in both Euclidean and hyperbolic spaces, while maintaining asymptotic complexity on par with Euclidean decision trees. 

The methodological innovation centers on selecting the appropriate decision boundary element to substitute for axis-parallel hyperplanes in the hyperbolic space context. 
Remarkably, homogeneous hyperplanes serve as highly effective building blocks, preserving continuity and convexity of subspaces at each partition and deviating from traditional approaches by offering straightforward expressions that avoid Riemannian optimization.
The trick of using single-axis rotations of the base plane further simplifies and speeds up computation.

%Overall, our algorithm is fast, simple, and stable.
%Using the hyperboloid model incurs less numerical instability due to floating-point issues, decision trees are fully interpretable with respect to their ultimate prediction behavior, and our code implements each predictor using the standard \sklrn\ API methods. 

\hyperdt\ and \hyperrf\ stand out for their speed, simplicity, and stability. 
The hyperboloid model uses a small, constant number of simple trigonometric expressions at each decision tree node, thereby minimizing numerical instability concerns arising from floating-point issues. 
Finally, the alignment of hyperboloid geometry to the tree-like structural features of hierarchy-oriented data manifests in single trees performing extraordinarily well, reducing the need for ensemble methods. 
Such single trees are faster to learn and use. More importantly, they offer full interpretability.

 We offer an implementation adhering to standard \sklrn\ API conventions, ensuring ease of use. Future research avenues can expand $\hyperrf$ with popular features such as gradient boosting, optimization enhancements (e.g., pruning), and additional applications for classification and regression within hyperbolic space.
 Additional optimizations can greatly improve both performance and usability e.g. through optimizing performance per the standards of \sklrn.
Furthermore, new advanced decision tree methods such as \citet{lin_generalized_2022}, \citet{mctavish_fast_2022}, and \citet{mazumder_quant-bnb_2022}, which are based on axis-parallel hyperplanes but apply non-greedy optimizations over multiple splits, can be reformulated in terms of dot-products and applied to homogeneous hyperplanes instead.

\subsubsection*{Acknowledgments}
We acknowledge the support of the NSF Graduate Research Fellowship under grant no. DGE-2036197 to Philippe Chlenski. We also thank Quentin Chu and Swati Negi for their early use and feedback on our software, which has been instrumental in its development.

\subsubsection*{Ethics statement}
This paper aims to advance the field of Machine Learning, conscious of its potential societal impacts and committed to adhering to the ICLR Code of Ethics. Although our research does not directly tackle sensitive ethical issues and we identify no specific societal consequences requiring individual emphasis, we recognize our work within the evolving landscape of machine learning ethics. Acknowledging that the ethical dimensions of machine learning are an area of ongoing exploration and debate, we commit to engaging responsibly with these broader considerations and adhering to ICLR guidelines to address any emerging concerns.

\subsubsection*{Reproducibility statement}
To ensure the reproducibility of our work, we have taken comprehensive steps detailed across our paper, its appendices, and the supplemental GitHub repository. The main text delineates the methodologies and experimental setups, with proofs and derivations of nontrivial mathematical insights in the Appendix. Our GitHub repository houses all of the code, data processing steps, and additional documentation that support the empirical results presented. The \texttt{README.md} file for our GitHub repo contains up-to-date, detailed information on which files and notebooks reproduce which parts of the paper and links to data that is not publically available. 

% In the GitHub repo, the file \texttt{hororf\_benchmarks.py} reproduces the benchmarking code used to generate Table~\ref{tab:accuracies}, Fig~\ref{fig:runtimes}, and the WordNet benchmarks in Appendix Subsection~\ref{appendix:wordnet}. Similarly, \texttt{maxdepth\_benchmarks.py} reproduces the experiments in Appendix \ref{appendix:scaling}. Code to carry out additional statistical tests for benchmarks can be found in \texttt{notebooks/25\_statistical\_testing\_benchmarks.ipynb}.

% The \texttt{notebooks/archive} subdirectory contains all Jupyter notebooks used to generate the rest of the results: in particular, 
% \begin{itemize}
%     \item \texttt{27\_good\_visualization.ipynb} reproduces Fig~\ref{fig:visualization},
%     \item \texttt{10\_fix\_midpoints.ipynb} reproduces Figs~\ref{fig:hyperboloid_intersections} and \ref{fig:angle_distances},
%     \item \texttt{31\_hsvm\_and\_logistic\_regression.ipynb} reproduces \ref{appendix:hsvm_hmlr},
%     \item \texttt{35\_all\_geometries.ipynb} reproduces \ref{appendix:other_geometries}, and 
%     \item \texttt{34\_ablations.ipynb} reproduces \ref{appendix:ablation}.
% \end{itemize}

\bibliography{references}

\begin{thebibliography}{40}
\providecommand{\natexlab}[1]{#1}
\providecommand{\url}[1]{\texttt{#1}}
\expandafter\ifx\csname urlstyle\endcsname\relax
  \providecommand{\doi}[1]{doi: #1}\else
  \providecommand{\doi}{doi: \begingroup \urlstyle{rm}\Url}\fi

\bibitem[Adamic \& Glance(2005)Adamic and Glance]{adamic_political_2005}
Lada~A. Adamic and Natalie Glance.
\newblock The political blogosphere and the 2004 {U}.{S}. election: divided
  they blog.
\newblock In \emph{Proceedings of the 3rd international workshop on {Link}
  discovery}, {LinkKDD} '05, pp.\  36--43, New York, NY, USA, August 2005.
  Association for Computing Machinery.
\newblock ISBN 978-1-59593-215-0.
\newblock \doi{10.1145/1134271.1134277}.
\newblock URL \url{https://doi.org/10.1145/1134271.1134277}.

\bibitem[Agibetov et~al.(2019)Agibetov, Dorffner, and
  Samwald]{agibetov_using_2019}
Asan Agibetov, Georg Dorffner, and Matthias Samwald.
\newblock Using hyperbolic large-margin classifiers for biological link
  prediction.
\newblock In \emph{Proceedings of the 5th {Workshop} on {Semantic} {Deep}
  {Learning} ({SemDeep}-5)}, pp.\  26--30, Macau, China, August 2019.
  Association for Computational Linguistics.
\newblock URL \url{https://aclanthology.org/W19-5805}.

\bibitem[Bdeir et~al.(2023)Bdeir, Schwethelm, and
  Landwehr]{bdeir_hyperbolic_2023}
Ahmad Bdeir, Kristian Schwethelm, and Niels Landwehr.
\newblock Hyperbolic {Geometry} in {Computer} {Vision}: {A} {Novel} {Framework}
  for {Convolutional} {Neural} {Networks}, March 2023.
\newblock URL \url{https://arxiv.org/abs/2303.15919v2}.

\bibitem[Breiman(2001)]{breiman_random_2001}
Leo Breiman.
\newblock Random forests.
\newblock \emph{Machine Learning}, 45\penalty0 (1):\penalty0 5--32, October
  2001.
\newblock ISSN 1573-0565.
\newblock \doi{10.1023/A:1010933404324}.
\newblock URL \url{https://doi.org/10.1023/A:1010933404324}.

\bibitem[Breiman(2017)]{breiman_classification_2017}
Leo Breiman.
\newblock \emph{Classification and {Regression} {Trees}}.
\newblock Routledge, New York, October 2017.
\newblock ISBN 978-1-315-13947-0.
\newblock \doi{10.1201/9781315139470}.

\bibitem[Chamberlain et~al.(2017)Chamberlain, Clough, and
  Deisenroth]{chamberlain_neural_2017}
Benjamin~Paul Chamberlain, James Clough, and Marc~Peter Deisenroth.
\newblock Neural {Embeddings} of {Graphs} in {Hyperbolic} {Space}, May 2017.
\newblock URL \url{http://arxiv.org/abs/1705.10359}.
\newblock arXiv:1705.10359 [cs, stat].

\bibitem[Chami et~al.(2019)Chami, Ying, Ré, and
  Leskovec]{chami_hyperbolic_2019}
Ines Chami, Rex Ying, Christopher Ré, and Jure Leskovec.
\newblock Hyperbolic {Graph} {Convolutional} {Neural} {Networks}, October 2019.
\newblock URL \url{http://arxiv.org/abs/1910.12933}.
\newblock arXiv:1910.12933 [cs, stat].

\bibitem[Chami et~al.(2020)Chami, Gu, Chatziafratis, and Ré]{chami_trees_2020}
Ines Chami, Albert Gu, Vaggos Chatziafratis, and Christopher Ré.
\newblock From {Trees} to {Continuous} {Embeddings} and {Back}: {Hyperbolic}
  {Hierarchical} {Clustering}, October 2020.
\newblock URL \url{https://arxiv.org/abs/2010.00402v1}.

\bibitem[Chami et~al.(2021)Chami, Gu, Nguyen, and Ré]{chami_horopca_2021}
Ines Chami, Albert Gu, Dat Nguyen, and Christopher Ré.
\newblock {HoroPCA}: {Hyperbolic} {Dimensionality} {Reduction} via
  {Horospherical} {Projections}, June 2021.
\newblock URL \url{http://arxiv.org/abs/2106.03306}.
\newblock arXiv:2106.03306 [cs].

\bibitem[Chen \& Guestrin(2016)Chen and Guestrin]{chen_xgboost_2016}
Tianqi Chen and Carlos Guestrin.
\newblock {XGBoost}: {A} {Scalable} {Tree} {Boosting} {System}.
\newblock In \emph{Proceedings of the 22nd {ACM} {SIGKDD} {International}
  {Conference} on {Knowledge} {Discovery} and {Data} {Mining}}, pp.\  785--794,
  August 2016.
\newblock \doi{10.1145/2939672.2939785}.
\newblock URL \url{http://arxiv.org/abs/1603.02754}.
\newblock arXiv:1603.02754 [cs].

\bibitem[Chen et~al.(2022)Chen, Han, Lin, Zhao, Liu, Li, Sun, and
  Zhou]{chen_fully_2022}
Weize Chen, Xu~Han, Yankai Lin, Hexu Zhao, Zhiyuan Liu, Peng Li, Maosong Sun,
  and Jie Zhou.
\newblock Fully {Hyperbolic} {Neural} {Networks}, March 2022.
\newblock URL \url{http://arxiv.org/abs/2105.14686}.
\newblock arXiv:2105.14686 [cs].

\bibitem[Cho et~al.(2018)Cho, DeMeo, Peng, and Berger]{cho_large-margin_2018}
Hyunghoon Cho, Benjamin DeMeo, Jian Peng, and Bonnie Berger.
\newblock Large-{Margin} {Classification} in {Hyperbolic} {Space}, June 2018.
\newblock URL \url{http://arxiv.org/abs/1806.00437}.
\newblock arXiv:1806.00437 [cs, stat].

\bibitem[Corso et~al.(2021)Corso, Ying, Pándy, Veličković, Leskovec, and
  Liò]{corso_neural_2021}
Gabriele Corso, Rex Ying, Michal Pándy, Petar Veličković, Jure Leskovec, and
  Pietro Liò.
\newblock Neural {Distance} {Embeddings} for {Biological} {Sequences}, October
  2021.
\newblock URL \url{http://arxiv.org/abs/2109.09740}.
\newblock arXiv:2109.09740 [cs, q-bio].

\bibitem[De~Sa et~al.(2018)De~Sa, Gu, Ré, and Sala]{de_sa_representation_2018}
Christopher De~Sa, Albert Gu, Christopher Ré, and Frederic Sala.
\newblock Representation {Tradeoffs} for {Hyperbolic} {Embeddings}, April 2018.
\newblock URL \url{http://arxiv.org/abs/1804.03329}.
\newblock arXiv:1804.03329 [cs, stat].

\bibitem[Desai et~al.(2023)Desai, Nickel, Rajpurohit, Johnson, and
  Vedantam]{desai_hyperbolic_2023}
Karan Desai, Maximilian Nickel, Tanmay Rajpurohit, Justin Johnson, and
  Ramakrishna Vedantam.
\newblock Hyperbolic {Image}-{Text} {Representations}, June 2023.
\newblock URL \url{http://arxiv.org/abs/2304.09172}.
\newblock arXiv:2304.09172 [cs].

\bibitem[Ding \& Regev(2021)Ding and Regev]{ding_deep_2021}
Jiarui Ding and Aviv Regev.
\newblock Deep generative model embedding of single-cell {RNA}-{Seq} profiles
  on hyperspheres and hyperbolic spaces.
\newblock \emph{Nature Communications}, 12\penalty0 (1):\penalty0 2554, May
  2021.
\newblock ISSN 2041-1723.
\newblock \doi{10.1038/s41467-021-22851-4}.
\newblock URL \url{https://www.nature.com/articles/s41467-021-22851-4}.
\newblock Number: 1 Publisher: Nature Publishing Group.

\bibitem[Doorenbos et~al.(2023)Doorenbos, Márquez-Neila, Sznitman, and
  Mettes]{doorenbos_hyperbolic_2023}
Lars Doorenbos, Pablo Márquez-Neila, Raphael Sznitman, and Pascal Mettes.
\newblock Hyperbolic {Random} {Forests}, August 2023.
\newblock URL \url{http://arxiv.org/abs/2308.13279}.
\newblock arXiv:2308.13279 [cs].

\bibitem[Fan et~al.(2023)Fan, Yang, and Vemuri]{fan_horospherical_2023}
Xiran Fan, Chun-Hao Yang, and Baba~C. Vemuri.
\newblock Horospherical {Decision} {Boundaries} for {Large} {Margin}
  {Classification} in {Hyperbolic} {Space}, June 2023.
\newblock URL \url{http://arxiv.org/abs/2302.06807}.
\newblock arXiv:2302.06807 [cs, stat].

\bibitem[Fellbaum(2010)]{fellbaum_wordnet_2010}
Christiane Fellbaum.
\newblock {WordNet}.
\newblock In Roberto Poli, Michael Healy, and Achilles Kameas (eds.),
  \emph{Theory and {Applications} of {Ontology}: {Computer} {Applications}},
  pp.\  231--243. Springer Netherlands, Dordrecht, 2010.
\newblock ISBN 978-90-481-8847-5.
\newblock \doi{10.1007/978-90-481-8847-5_10}.
\newblock URL \url{https://doi.org/10.1007/978-90-481-8847-5_10}.

\bibitem[Ganea et~al.(2018)Ganea, Bécigneul, and
  Hofmann]{ganea_hyperbolic_2018}
Octavian-Eugen Ganea, Gary Bécigneul, and Thomas Hofmann.
\newblock Hyperbolic {Entailment} {Cones} for {Learning} {Hierarchical}
  {Embeddings}, June 2018.
\newblock URL \url{http://arxiv.org/abs/1804.01882}.
\newblock arXiv:1804.01882 [cs, stat].

\bibitem[Gu et~al.(2019)Gu, Sala, Gunel, and Re]{gu_learning_2019}
Albert Gu, Frederic Sala, Beliz Gunel, and Christopher Re.
\newblock Learning mixed-curvature representations in products of model spaces.
\newblock 2019.

\bibitem[Gulcehre et~al.(2018)Gulcehre, Denil, Malinowski, Razavi, Pascanu,
  Hermann, Battaglia, Bapst, Raposo, Santoro, and
  de~Freitas]{gulcehre_hyperbolic_2018}
Caglar Gulcehre, Misha Denil, Mateusz Malinowski, Ali Razavi, Razvan Pascanu,
  Karl~Moritz Hermann, Peter Battaglia, Victor Bapst, David Raposo, Adam
  Santoro, and Nando de~Freitas.
\newblock Hyperbolic {Attention} {Networks}, May 2018.
\newblock URL \url{http://arxiv.org/abs/1805.09786}.
\newblock arXiv:1805.09786 [cs].

\bibitem[Hughes et~al.(2004)Hughes, Hyun, and
  Liberles]{hughes_visualising_2004}
Timothy Hughes, Young Hyun, and David~A. Liberles.
\newblock Visualising very large phylogenetic trees in three dimensional
  hyperbolic space.
\newblock \emph{BMC Bioinformatics}, 5\penalty0 (1):\penalty0 48, April 2004.
\newblock ISSN 1471-2105.
\newblock \doi{10.1186/1471-2105-5-48}.
\newblock URL \url{https://doi.org/10.1186/1471-2105-5-48}.

\bibitem[Jiang et~al.(2022)Jiang, Tabaghi, and Mirarab]{jiang_learning_2022}
Yueyu Jiang, Puoya Tabaghi, and Siavash Mirarab.
\newblock Learning {Hyperbolic} {Embedding} for {Phylogenetic} {Tree}
  {Placement} and {Updates}.
\newblock \emph{Biology}, 11\penalty0 (9):\penalty0 1256, September 2022.
\newblock ISSN 2079-7737.
\newblock \doi{10.3390/biology11091256}.
\newblock URL \url{https://www.mdpi.com/2079-7737/11/9/1256}.
\newblock Number: 9 Publisher: Multidisciplinary Digital Publishing Institute.

\bibitem[Lin et~al.(2022)Lin, Zhong, Hu, Rudin, and
  Seltzer]{lin_generalized_2022}
Jimmy Lin, Chudi Zhong, Diane Hu, Cynthia Rudin, and Margo Seltzer.
\newblock Generalized and {Scalable} {Optimal} {Sparse} {Decision} {Trees},
  November 2022.
\newblock URL \url{http://arxiv.org/abs/2006.08690}.
\newblock arXiv:2006.08690 [cs, stat].

\bibitem[Marconi et~al.(2020)Marconi, Rosasco, and
  Ciliberto]{marconi_hyperbolic_2020}
Gian~Maria Marconi, Lorenzo Rosasco, and Carlo Ciliberto.
\newblock Hyperbolic {Manifold} {Regression}, May 2020.
\newblock URL \url{http://arxiv.org/abs/2005.13885}.
\newblock arXiv:2005.13885 [cs, stat].

\bibitem[Mazumder et~al.(2022)Mazumder, Meng, and
  Wang]{mazumder_quant-bnb_2022}
Rahul Mazumder, Xiang Meng, and Haoyue Wang.
\newblock Quant-{BnB}: {A} {Scalable} {Branch}-and-{Bound} {Method} for
  {Optimal} {Decision} {Trees} with {Continuous} {Features}, June 2022.
\newblock URL \url{http://arxiv.org/abs/2206.11844}.
\newblock arXiv:2206.11844 [cs].

\bibitem[McDonald et~al.(2018)McDonald, Hyde, Debelius, Morton, Gonzalez,
  Ackermann, Aksenov, Behsaz, Brennan, Chen, DeRight~Goldasich, Dorrestein,
  Dunn, Fahimipour, Gaffney, Gilbert, Gogul, Green, Hugenholtz, Humphrey,
  Huttenhower, Jackson, Janssen, Jeste, Jiang, Kelley, Knights, Kosciolek,
  Ladau, Leach, Marotz, Meleshko, Melnik, Metcalf, Mohimani, Montassier,
  Navas-Molina, Nguyen, Peddada, Pevzner, Pollard, Rahnavard, Robbins-Pianka,
  Sangwan, Shorenstein, Smarr, Song, Spector, Swafford, Thackray, Thompson,
  Tripathi, Vázquez-Baeza, Vrbanac, Wischmeyer, Wolfe, Zhu, {American Gut
  Consortium}, and Knight]{mcdonald_american_2018}
Daniel McDonald, Embriette Hyde, Justine~W. Debelius, James~T. Morton, Antonio
  Gonzalez, Gail Ackermann, Alexander~A. Aksenov, Bahar Behsaz, Caitriona
  Brennan, Yingfeng Chen, Lindsay DeRight~Goldasich, Pieter~C. Dorrestein,
  Robert~R. Dunn, Ashkaan~K. Fahimipour, James Gaffney, Jack~A. Gilbert, Grant
  Gogul, Jessica~L. Green, Philip Hugenholtz, Greg Humphrey, Curtis
  Huttenhower, Matthew~A. Jackson, Stefan Janssen, Dilip~V. Jeste, Lingjing
  Jiang, Scott~T. Kelley, Dan Knights, Tomasz Kosciolek, Joshua Ladau, Jeff
  Leach, Clarisse Marotz, Dmitry Meleshko, Alexey~V. Melnik, Jessica~L.
  Metcalf, Hosein Mohimani, Emmanuel Montassier, Jose Navas-Molina, Tanya~T.
  Nguyen, Shyamal Peddada, Pavel Pevzner, Katherine~S. Pollard, Gholamali
  Rahnavard, Adam Robbins-Pianka, Naseer Sangwan, Joshua Shorenstein, Larry
  Smarr, Se~Jin Song, Timothy Spector, Austin~D. Swafford, Varykina~G.
  Thackray, Luke~R. Thompson, Anupriya Tripathi, Yoshiki Vázquez-Baeza, Alison
  Vrbanac, Paul Wischmeyer, Elaine Wolfe, Qiyun Zhu, {American Gut Consortium},
  and Rob Knight.
\newblock American {Gut}: an {Open} {Platform} for {Citizen} {Science}
  {Microbiome} {Research}.
\newblock \emph{mSystems}, 3\penalty0 (3):\penalty0 e00031--18, 2018.
\newblock ISSN 2379-5077.
\newblock \doi{10.1128/mSystems.00031-18}.

\bibitem[McDonald et~al.(2023)McDonald, Jiang, Balaban, Cantrell, Zhu,
  Gonzalez, Morton, Nicolaou, Parks, Karst, Albertsen, Hugenholtz, DeSantis,
  Song, Bartko, Havulinna, Jousilahti, Cheng, Inouye, Niiranen, Jain, Salomaa,
  Lahti, Mirarab, and Knight]{mcdonald_greengenes2_2023}
Daniel McDonald, Yueyu Jiang, Metin Balaban, Kalen Cantrell, Qiyun Zhu, Antonio
  Gonzalez, James~T. Morton, Giorgia Nicolaou, Donovan~H. Parks, Søren~M.
  Karst, Mads Albertsen, Philip Hugenholtz, Todd DeSantis, Se~Jin Song, Andrew
  Bartko, Aki~S. Havulinna, Pekka Jousilahti, Susan Cheng, Michael Inouye,
  Teemu Niiranen, Mohit Jain, Veikko Salomaa, Leo Lahti, Siavash Mirarab, and
  Rob Knight.
\newblock Greengenes2 unifies microbial data in a single reference tree.
\newblock \emph{Nature Biotechnology}, pp.\  1--4, July 2023.
\newblock ISSN 1546-1696.
\newblock \doi{10.1038/s41587-023-01845-1}.
\newblock URL \url{https://www.nature.com/articles/s41587-023-01845-1}.
\newblock Publisher: Nature Publishing Group.

\bibitem[McTavish et~al.(2022)McTavish, Zhong, Achermann, Karimalis, Chen,
  Rudin, and Seltzer]{mctavish_fast_2022}
Hayden McTavish, Chudi Zhong, Reto Achermann, Ilias Karimalis, Jacques Chen,
  Cynthia Rudin, and Margo Seltzer.
\newblock Fast {Sparse} {Decision} {Tree} {Optimization} via {Reference}
  {Ensembles}, July 2022.
\newblock URL \url{http://arxiv.org/abs/2112.00798}.
\newblock arXiv:2112.00798 [cs].

\bibitem[Miolane et~al.(2018)Miolane, Mathe, Donnat, Jorda, and
  Pennec]{miolane_geomstats_2018}
Nina Miolane, Johan Mathe, Claire Donnat, Mikael Jorda, and Xavier Pennec.
\newblock geomstats: a {Python} {Package} for {Riemannian} {Geometry} in
  {Machine} {Learning}, May 2018.
\newblock URL \url{https://arxiv.org/abs/1805.08308v2}.

\bibitem[Nagano et~al.(2019)Nagano, Yamaguchi, Fujita, and
  Koyama]{nagano_wrapped_2019}
Yoshihiro Nagano, Shoichiro Yamaguchi, Yasuhiro Fujita, and Masanori Koyama.
\newblock A {Wrapped} {Normal} {Distribution} on {Hyperbolic} {Space} for
  {Gradient}-{Based} {Learning}, May 2019.
\newblock URL \url{http://arxiv.org/abs/1902.02992}.
\newblock arXiv:1902.02992 [cs, stat].

\bibitem[Nickel \& Kiela(2017)Nickel and Kiela]{nickel_poincare_2017}
Maximilian Nickel and Douwe Kiela.
\newblock Poincar{\textbackslash}'e {Embeddings} for {Learning} {Hierarchical}
  {Representations}, May 2017.
\newblock URL \url{http://arxiv.org/abs/1705.08039}.
\newblock arXiv:1705.08039 [cs, stat].

\bibitem[Nickel \& Kiela(2018)Nickel and Kiela]{nickel_learning_2018}
Maximilian Nickel and Douwe Kiela.
\newblock Learning {Continuous} {Hierarchies} in the {Lorentz} {Model} of
  {Hyperbolic} {Geometry}, July 2018.
\newblock URL \url{http://arxiv.org/abs/1806.03417}.
\newblock arXiv:1806.03417 [cs, stat].

\bibitem[Pedregosa et~al.(2011)Pedregosa, Varoquaux, Gramfort, Michel, Thirion,
  Grisel, Blondel, Prettenhofer, Weiss, Dubourg, Vanderplas, Passos,
  Cournapeau, Brucher, Perrot, and Duchesnay]{pedregosa_scikit-learn_2011}
Fabian Pedregosa, Gaël Varoquaux, Alexandre Gramfort, Vincent Michel, Bertrand
  Thirion, Olivier Grisel, Mathieu Blondel, Peter Prettenhofer, Ron Weiss,
  Vincent Dubourg, Jake Vanderplas, Alexandre Passos, David Cournapeau,
  Matthieu Brucher, Matthieu Perrot, and Édouard Duchesnay.
\newblock Scikit-learn: {Machine} {Learning} in {Python}.
\newblock \emph{Journal of Machine Learning Research}, 12\penalty0
  (85):\penalty0 2825--2830, 2011.
\newblock ISSN 1533-7928.
\newblock URL \url{http://jmlr.org/papers/v12/pedregosa11a.html}.

\bibitem[Radford et~al.(2021)Radford, Kim, Hallacy, Ramesh, Goh, Agarwal,
  Sastry, Askell, Mishkin, Clark, Krueger, and
  Sutskever]{radford_learning_2021}
Alec Radford, Jong~Wook Kim, Chris Hallacy, Aditya Ramesh, Gabriel Goh,
  Sandhini Agarwal, Girish Sastry, Amanda Askell, Pamela Mishkin, Jack Clark,
  Gretchen Krueger, and Ilya Sutskever.
\newblock Learning {Transferable} {Visual} {Models} {From} {Natural} {Language}
  {Supervision}, February 2021.
\newblock URL \url{http://arxiv.org/abs/2103.00020}.
\newblock arXiv:2103.00020 [cs].

\bibitem[Sarkar(2012)]{van_kreveld_low_2012}
Rik Sarkar.
\newblock Low {Distortion} {Delaunay} {Embedding} of {Trees} in {Hyperbolic}
  {Plane}.
\newblock In Marc Van~Kreveld and Bettina Speckmann (eds.), \emph{Graph
  {Drawing}}, volume 7034, pp.\  355--366. Springer Berlin Heidelberg, Berlin,
  Heidelberg, 2012.
\newblock ISBN 978-3-642-25877-0 978-3-642-25878-7.
\newblock \doi{10.1007/978-3-642-25878-7_34}.
\newblock URL \url{http://link.springer.com/10.1007/978-3-642-25878-7_34}.
\newblock Series Title: Lecture Notes in Computer Science.

\bibitem[Tay et~al.(2018)Tay, Tuan, and Hui]{tay_hyperbolic_2018}
Yi~Tay, Luu~Anh Tuan, and Siu~Cheung Hui.
\newblock Hyperbolic {Representation} {Learning} for {Fast} and {Efficient}
  {Neural} {Question} {Answering}.
\newblock In \emph{Proceedings of the {Eleventh} {ACM} {International}
  {Conference} on {Web} {Search} and {Data} {Mining}}, {WSDM} '18, pp.\
  583--591, New York, NY, USA, February 2018. Association for Computing
  Machinery.
\newblock ISBN 978-1-4503-5581-0.
\newblock \doi{10.1145/3159652.3159664}.
\newblock URL \url{https://doi.org/10.1145/3159652.3159664}.

\bibitem[Tifrea et~al.(2018)Tifrea, Bécigneul, and
  Ganea]{tifrea_poincare_2018}
Alexandru Tifrea, Gary Bécigneul, and Octavian-Eugen Ganea.
\newblock Poincar{\textbackslash}'e {GloVe}: {Hyperbolic} {Word} {Embeddings},
  November 2018.
\newblock URL \url{http://arxiv.org/abs/1810.06546}.
\newblock arXiv:1810.06546 [cs].

\bibitem[van Spengler et~al.(2023)van Spengler, Wirth, and
  Mettes]{van_spengler_hypll_2023}
Max van Spengler, Philipp Wirth, and Pascal Mettes.
\newblock {HypLL}: {The} {Hyperbolic} {Learning} {Library}, August 2023.
\newblock URL \url{http://arxiv.org/abs/2306.06154}.
\newblock arXiv:2306.06154 [cs].

\end{thebibliography}
\bibliographystyle{template/iclr2024_conference}

\appendix
\section{Appendix}

\subsection{Conversion between Hyperboloid and Poincar\'e models}
\label{appendix:conversion}
% Converting between the $D$-dimensional hyperboloid $\mathbb{H}^D$ and Poincar\'e disk $\mathbb{P}^D$ is simple when $K=1$.
Letting $\mathbf{x_P}$ be a point in $\mathbb{P}^{D,K}$ and $\mathbf{x_H}$ be its equivalent in $\hdk$,
\begin{align}
    x_{P, i} &= \frac{x_{H,i}}{\sqrt{K} + x_{H,0}} \label{eq:hyperboloid_to_poincare}\\
    x_{H, 0} &= \sqrt{K}\cdot\frac{1 + \|x_P\|^2_2}{1 - \|x_P\|^2_2}\\
    x_{H, i} &= \sqrt{K}\cdot\frac{2 x_{P,i}}{1 - \|x_P\|_2^2}.
\end{align}

\subsection{Geodesic submanifold details}
\label{appendix:visualization}
In this section, we will give a full parameterization of the geodesic submanifolds created by intersecting the decision hyperplanes learned by \hyperdt\ and \hyperrf\ with $\hdk$. 
% We proceed in four phases:
We enumerate the basis vectors of our decision hyperplanes using a formalism conducive to the following steps, calculate a scaling factor that ensures a basis vector reaches the surface of the manifold, construct a 1-dimensional geodesic arc from the basis vectors of a decision hyperplane, and finally recursively extend this arc to higher dimensions, culminating in a full $(D-1)$-dimensional geodesic submanifold.

To reduce the complexity of notation necessitated by indexing over dimensions, we will assume without loss of generality that the decision hyperplane's normal vector is nonzero in dimensions 0 and 1.
%We will further simplify our notation by requiring that $d$-dimensional geodesics be nonzero in the first $d+1$ dimensions: therefore, we will choose dimension 2 to parameterize a one-dimensional geodesic, 3 to turn it into a 2-dimensional submanifold, and so on. 
We extend this convention to geodesic submanifolds by putting nonzero dimensions first: therefore, we first parameterize an arc along dimension 2, then turn it into a 2-dimensional submanifold along dimension 3, and so on. 
In general, a $(d \leq D)$-dimensional submanifold will be nonzero in the first $d+1$ ambient dimensions.

\subsubsection{Basis vectors of decision hyperplanes}
\label{appendix:basis_vectors}
Let $\mathbf{P}(\theta)$ be a $D$-dimensional decision hyperplane learned by \hyperdt.
By our assumption above and Equation~\ref{eq:normal}, the normal vector $\mathbf{n}(1, \theta)$ of $\mathbf{P}(\theta)$ is nonzero only in dimensions 0 and 1.
% It can be made by inclining $x_0=0$ by $\theta$ degrees in the manner of the decision hyperplanes, with the normal vector $\mathbf{n}(1, \theta)$ as given in Equation~\ref{eq:normal}. By our assumption, this plane is inclined in the first dimension. 
Therefore $\mathbf{P}(\theta)$ has $D$ basis vectors: 
%$\mathbf{v} = \langle \sin(\theta), \cos(\theta), \ldots 0 \rangle$, and a set of one-hot vectors $\mathbf{u^d}$ where $u^d_{d} = 1$ for each $2 \leq d \leq D$.
\begin{align}
    \mathbf{v^0} &= \langle \sin(\theta),\ \cos(\theta),\ 0,\ \ldots \rangle\\
    \label{eq:ud_vector}
    \mathbf{u^d} &= \langle 0,\ \ldots,\ u_d^d=1,\  \ldots,\ 0 \rangle,\ 2 \leq d \leq D
\end{align}
The $\mathbf{u^d}$ vectors are standard basis vectors which are 1 in dimension $d$ and 0 elsewhere.

% \subsubsection{Computing $\alpha$}
\subsubsection{Computing vector scale}
\label{appendix:alpha}
% Our 2-plane has two basis vectors, $\mathbf{v}$ with $v_0 = \sin(\theta)$ and $v_d = \cos(\theta)$ and $\mathbf{u}$ has $u_{d'} = 1$ for some $d'$ which is neither 0 nor $d$.

% Let $\mathbf{v}$ be the basis vector in dimensions 0 and 1 as described in Section~\ref{appendix:basis_vectors}.
For what scaling factor $\alpha$ does $\alpha\mathbf{v^0}$ lie on the manifold?
We know that $\alpha\mathbf{v^0}$ will always be zero in dimensions 2 through $D$, effectively reducing this to a simple 2-dimensional problem on $\mathbb{H}^{2,K}$:
% For simplicity, we can restrict ourselves to the case where only dimensions 0 and 1 are nonzero, effectively restricting ourselves to $\mathbb{H}^{2,K}$.

% By our assumption that other dimensions are zero, $\hdk$ has a simple 2-dimensional form, which we can solve for $\alpha$:
\begin{align}
    x_1^2 - x_0^2 &= -1/K\\
    x_1^2 &= -1/K + x_0^2\\
    \alpha^2\cos^2(\theta) &= -1/K + \alpha^2\sin^2(\theta)\\
    \alpha^2(\cos^2(\theta) - \sin^2(\theta))  &= -1/K\\
    K\alpha^2 &= \frac{-1}{\cos^2(\theta) - \sin^2(\theta)} \label{eq:pre_double_angle}\\
    K\alpha^2 &= \frac{-1}{\cos(2\theta)} \label{eq:post_double_angle}\\
    % K\alpha^2 &= -\sec(2\theta)\\
    % \sqrt{K}\alpha & = \sqrt{-\sec(2\theta)}\\
    \alpha &= \sqrt{\frac{{-\sec(2\theta)}}{K}}
\end{align}
% \begin{align}
%     x_d^2 - x_0^2 &= -1\\
%     x_d^2 &= -1 + x_0^2\\
%     \alpha^2\cos^2(\theta) &= -1 + \alpha^2\sin^2(\theta)\\
%     \alpha^2(\cos^2(\theta) - \sin^2(\theta))  &= -1\\
%     \alpha^2 &= \frac{-1}{\cos^2(\theta) - \sin^2(\theta)} \label{eq:pre_double_angle}\\
%     \alpha^2 &= \frac{-1}{\cos(2\theta)} \label{eq:post_double_angle}\\
%     \alpha^2 &= -\sec(2\theta)\\
%     \alpha & = \sqrt{-\sec(2\theta)}
% \end{align}

\begin{figure}[h]
\caption{Rescaling basis vector $\mathbf{v^0} = \langle \sin(\theta), \cos(\theta) \rangle$ by $\alpha(\theta, 1) = \sqrt{-\sec(2\theta)}$ produces a point on $\mathbb{H}^{1,1}$ for all $\theta$ values between $\pi/4$ and $3\pi/4$.}
\label{fig:hyperboloid_intersections}
\begin{center}
\includegraphics[width=0.8\textwidth]{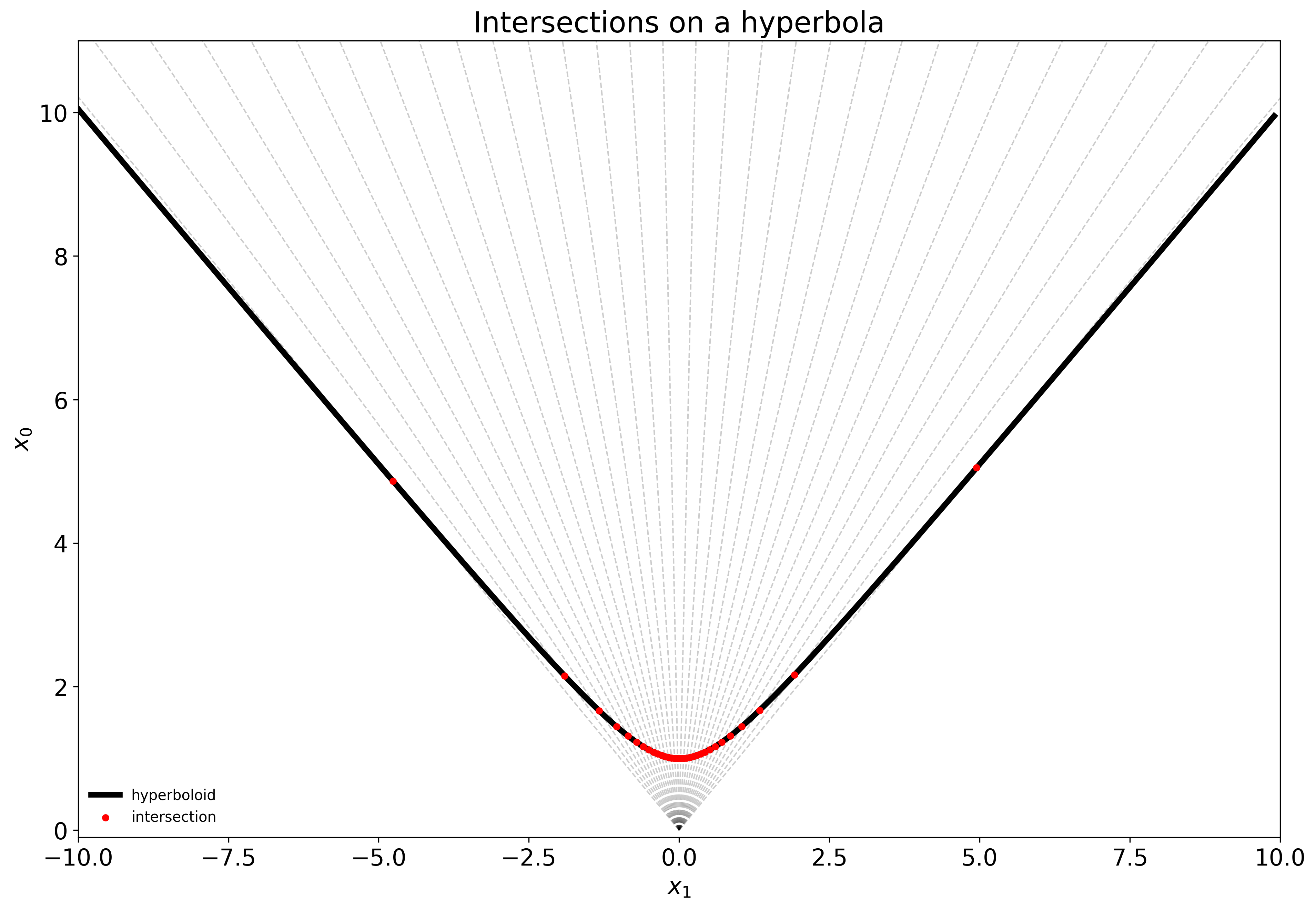}
\end{center}
\end{figure}

The transition from Equation \ref{eq:pre_double_angle} to Equation \ref{eq:post_double_angle} is due to the double-angle formula.
See Figure \ref{fig:hyperboloid_intersections} for a visual demonstration that rescaling $\mathbf{v^0}$ by $\alpha$ works for the full range of $\theta$ values in one dimension. To extend this insight to arbitrary angles and curvatures, we define an $\alpha(\theta, K)$ function
\begin{equation}
    \label{eq:alpha}
    \alpha(\theta, K) 
    = \sqrt{\frac{{-\sec(2\theta)}}{K}}
    = \frac{\sqrt{-\sec(2\theta)}}{\sqrt{K}}.
\end{equation}

\subsubsection{Geodesic arcs}
\label{appendix:geodesic_arc}
As shown in \citet{chami_hyperbolic_2019}, a geodesic arc in $\hdk$ can be characterized as $\cosh(t)\mathbf{v}^* + \sinh(t)\mathbf{u}^*$ for any pair of vectors $(\mathbf{u}^*, \mathbf{v}^*)$ where
\begin{align}
    \label{eq:geodesic_condition_1}
    \langle \mathbf{u}^*, \mathbf{u}^* \rangle_{\mathcal{L}} &= 1/K\\
    \label{eq:geodesic_condition_2}
    \langle \mathbf{v}^*, \mathbf{v}^* \rangle_{\mathcal{L}} &= -1/K\\
    \label{eq:geodesic_condition_3}
    \langle \mathbf{u}^*, \mathbf{v}^* \rangle_{\mathcal{L}} &= 0.
\end{align}

% Since $\alpha(\theta, K)\mathbf{v}$ is on the hyperboloid, $\langle \alpha(\theta, K)\mathbf{v}, \alpha(\theta,K)\mathbf{v} \rangle_\mathcal{L} = -1/K$.
All $\mathbf{u^d}$ vectors described in Equation~\ref{eq:ud_vector}, being purely spacelike, satisfy Equation~\ref{eq:geodesic_condition_1} if rescaled by a factor of $\sqrt{K}$ so that their norms are $1/K$.
We arbitrarily choose to use $\mathbf{u^2}/\sqrt{K}$ in our paramterization. 
Since $\alpha(\theta, K)\mathbf{v^0}$ lies on $\hdk$, it satisfies Equation~\ref{eq:geodesic_condition_2}.
Since $\mathbf{v^0}$ and $\mathbf{u^d}$ are disjoint in their nonzero dimensions, they trivially satisfy Equation~\ref{eq:geodesic_condition_3} for any scaling factors.
% Each of the other basis vectors $\mathbf{u}_{d'}$, being orthogonal to $\mathbf{v}$ and purely spacelike, satisfies the other two conditions if rescaled by $1/\sqrt{K}$ to the correct norm. We call this rescaled vector $\mathbf{u}(d', K)$.
Letting $t$ vary freely, the geodesic is given by
\begin{align}
    \label{eq:geodesic_1d}
    \mathbf{g^1}(\theta, K,  t) 
    &= \cosh(t)\cdot\alpha(\theta, K)\cdot\mathbf{v^0} + \sinh(t)\mathbf{u^2}/\sqrt{K}\\
    &= \langle 
    \cosh(t)\cdot\alpha(\theta, K)\cdot\sin(\theta),\ 
    \cosh(t)\cdot\alpha(\theta, K)\cdot\cos(\theta),\ 
    \sinh(t)/\sqrt{K},\ 
    0,
    \ldots
    \rangle.
\end{align}
% \begin{align}
%     \label{eq:geodesic_arc}
%     g_0(K, \theta, t) &= \cosh(t)\ \alpha(\theta,K)\sin(\theta)\\
%     g_1(K, \theta, t) &= \cosh(t)\ \alpha(\theta,K)\cos(\theta)\\
%     g_2(K, \theta, t) &= \sinh(t) / \sqrt{K}\\
%     g_{x}(K, \theta, t) &= 0 \text{ for } x \notin \{0, d, d'\}
% \end{align}
% For any $t \in \mathbb{R}$, the described point will be on the same geodesic arc.
% Extending this geodesic arc out to infinity partitions the hyperboloid.
The full geodesic arc over all possible values of $t$ is given by
\begin{equation}
    \label{eq:geodesic_arc}
    \mathbf{G^1}(\theta, K) = \left\{\mathbf{g^1}(\theta, K, t)\ :\ t \in \mathbb{R} \right\}.
\end{equation}

\subsubsection{Geodesic submanifolds}
\label{appendix:geodesic_submanifold}
%Let $\mathbf{G}$ be some geodesic submanifold---for instance, $\mathbf{G}(\theta, K)$ from Equation~\ref{eq:geodesic_arc}.
For didactic purposes, we first extend the geodesic arc $\mathbf{G^1}$ to a 2-dimensional submanifold $\mathbf{G^2}$.

Since any point in $\mathbf{G^1}(\theta, K)$ is on $\hdk$, it has Minkowski norm $-1/K$ and therefore satisfies Condition~\ref{eq:geodesic_condition_2}.

By our convention, we use $\mathbf{u^d}$ vectors sequentially to construct geodesics.
Therefore, if $\mathbf{G^d}$ is $d$-dimensional, then all $\mathbf{u^{d'}}$ for $d+2 \leq d' \leq D$, being unused in the construction of the geodesic, remain orthogonal to all $\mathbf{v^{d}} \in \mathbf{G^d}$ and continue to satisfy Condition~\ref{eq:geodesic_condition_3}.
In particular, $\mathbf{u^3}$ is the smallest unused $\mathbf{u^d}$ vector.
Being spacelike, all $\mathbf{u^{d'}}$ continue to satisfy Condition~\ref{eq:geodesic_condition_1}.

For our next geodesic, we apply Equation~\ref{eq:geodesic_1d} recursively to any $\mathbf{v^1} \in \mathbf{G^1}(\theta, K)$ and $\mathbf{u^3}/\sqrt{K}$: 
% Thus, the geodesic arc can be extended to a new dimension by rescaling the new $\mathbf{u'}$ by $1/\sqrt{K}$ and applying Equation~\ref{eq:geodesic_arc} once again. 
% For $\mathbf{v}' \in \mathbf{G}(\theta, K)$, we can define $\mathbf{g}'(K, \theta, t')$ recursively:
% Letting $\mathbf{v}'$ be $\mathbf{g}(t)$ for some $t$, we can define $\mathbf{g}'(t')$ recursively:
\begin{align}
    \mathbf{g^2}(\theta, K, t, t') &= \cosh(t')\mathbf{v^1} + \sinh(t')\mathbf{u^3}/\sqrt{K}\\
    &= \cosh(t')\mathbf{g^1}(\theta, K, t) + \sinh(t')\mathbf{u^3}/\sqrt{K}\\
    &= \cosh(t')(\cosh(t)\mathbf{v^0} + \sinh(t)\mathbf{u^2}) + \sinh(t')\mathbf{u^3}/\sqrt{K}
\end{align}
The geodesic submanifold created by intersecting the 3-plane with basis vectors $\{ \mathbf{v^0}, \mathbf{u^2}, \mathbf{u^3}\}$ with $\hdk$ corresponds to the set of all values of $\mathbf{g^2}$ for $(t, t') \in \mathbb{R}^2$:
\begin{equation}
    \mathbf{G^2}(\theta, K) = \{ \mathbf{g^2}(\theta, K, t, t') \ :\ (t, t') \in \mathbb{R}^2 \}
\end{equation}
% This process can be repeated for the $D-3$ remaining basis vectors, resulting in a parameterization over $\mathbf{t} \in \mathbb{R}^{D-1}$.

Using the remaining $\mathbf{u^d}$ vectors in ascending order from $d=3$ to $D$, we can recursively parameterize the full geodesic submanifold resulting from intersecting $\mathbf{P}(\theta)$ with $\hdk$:
\begin{align}
    % \mathbf{G}^n(K,\theta, 1) = \{\sinh(t)\mathbf{u}(n, K) + \cosh(t)\mathbf{v} : \mathbf{v} \in \mathbf{G}^{n-1}(K, \theta, 1, n-1),\ t \in \mathbb{R} \}
    \mathbf{G^{d}}(\theta, K) = \{\sinh(t)\mathbf{u^{d+1}} / \sqrt{K} + \cosh(t)\mathbf{v^{d-1}} : \mathbf{v^{d-1}} \in \mathbf{G^{d-1}}(\theta, K),\ t \in \mathbb{R} \}
\end{align}
% Using this definition, $\mathbf{G}^{D-1}(\theta, K)$ contains all points on the geodesic submanifold induced by intersection a $\mathbf{P}(\theta,1)$ decision plane inclined by $\theta$ degrees from $x_0=0$ along the first dimension, with $\hdk$.
% It is parameterized by all vectors $\mathbf{t} \in \mathbb{R}^D$.

% We finally extend this to the intersection of a full $D$-dimensional hyperplane with $\hdk$.
% Without loss of generality, let $d = 1$ and assume we select our $\mathbf{u}$ vectors in ascending order from 2 to $D$. We let 

% The full geodesic submanifold is then
% \begin{equation}
%     \mathbf{G^*}(K, \theta, d) = \left\{\mathbf{g^*}(K, \theta, d, d', t)\ :\ t \in \mathbb{R}^{D-1} \right\}
% \end{equation}

% In three dimensions, this looks like
% \begin{align}
%     g_0(t) &= \cosh(t')\cosh(t)\ \alpha\sin(\theta)\\
%     g_d(t) &= \cosh(t')\cosh(t)\ \alpha\cos(\theta)\\
%     g_{d'}(t) &= \cosh(t')\sinh(t) / \sqrt{K}\\
%     g_{d''}(t) &= \sinh(t') / \sqrt{K}
% \end{align}

\subsubsection{Visualization-specific details}
\paragraph{Visualization assumptions.}
The first step to plotting a learned decision boundary is to find closed form equations for the intersection between the hyperboloid and the plane.
To this end, we make a number of simplifying assumptions.
First of all, we restrict ourselves to the hyperboloid $\mathbb{H}^{2,1}$. 
% We also assume that our plane $\mathbf{P}$ can be parameterized by a single dimension $d$ and angle of inclination $\theta$.
For decision tree visualization, hyperplanes can be inclined along dimensions 1 or 2; therefore, we cannot assume that our first dimension contains the split.
Instead, we parameterize our plane as $\mathbf{P}(\theta, d)$.
The details of geodesics are the same as Equation~\ref{eq:geodesic_arc}, but dimensions 1 and 2 may be exchanged when $d=2$.
We also assume that the plane actually intersects $\mathbb{H}^{2,1}$, meaning $\pi/4 < \theta < 3\pi/4$.

\paragraph{Poincar\'e disk projection.}
Since $\mathbb{H}^{2,1}$ is actually a 3-dimensional object for visualization purposes, it is easier to visualize as a point on the Poincar\'e disk $\mathbb{P}^{2,1}$.
Thus, we convert coordinates in $\mathbb{H}^{2,1}$ to $\mathbb{P}^{2,1}$ using Equation \ref{eq:hyperboloid_to_poincare}.
For geodesics, we sample 1,000 points uniformly from $(-10, 10)$ and convert these: this is sufficient to draw a smooth arc on $\mathbb{P}^{2,1}$.

\paragraph{Subspace coloring.}
For better visualizations, it is also necessary to partition $\mathbb{P}^{2,1}$ so that:
\begin{enumerate}
    \item Decision boundaries are only rendered in the correct subtrees. For instance, if a boundary operates in the left subtree of a higher split, then it should only be drawn in the half of $\mathbb{P}^{2,1}$ where the left subtree is actually active.
    \item The space is partitioned fully by the leaves of the decision tree, and can therefore be colored according to the majority class at each leaf node.
\end{enumerate}

To do this, we recursively feed in a mask at each plotting iteration. 
This mask turns off plotting for inactive regions of the Poincar\'e disk.
At the leaf level, every point on the Poincar\'e disk is active in only one mask, and therefore can be used to plot majority classes.

\subsection{Midpoint angles}
\label{appendix:midpoints}
Now we consider how to find $\theta_m$, the midpoint between two angles $\theta_1$ and $\theta_2$. One option is to take the midpoint naively by taking the average of two angles:

\begin{equation}
\theta_{m, \text{naive}} = \frac{\theta_1 + \theta_2}{2}.
\end{equation}

\begin{figure}[h]
\caption{A plot of function $\delta(\pi / 4 + .01, \theta)$ as $\theta$ varies from $\pi/4$ to $3\pi/4$. This plot reveals the nonlinearity of the angle distance function.}
\label{fig:angle_distances}
\begin{center}
%\framebox[4.0in]{$\;$}
% \fbox{\rule[-.5cm]{0cm}{4cm} \rule[-.5cm]{4cm}{0cm}}
\includegraphics[width=\textwidth]{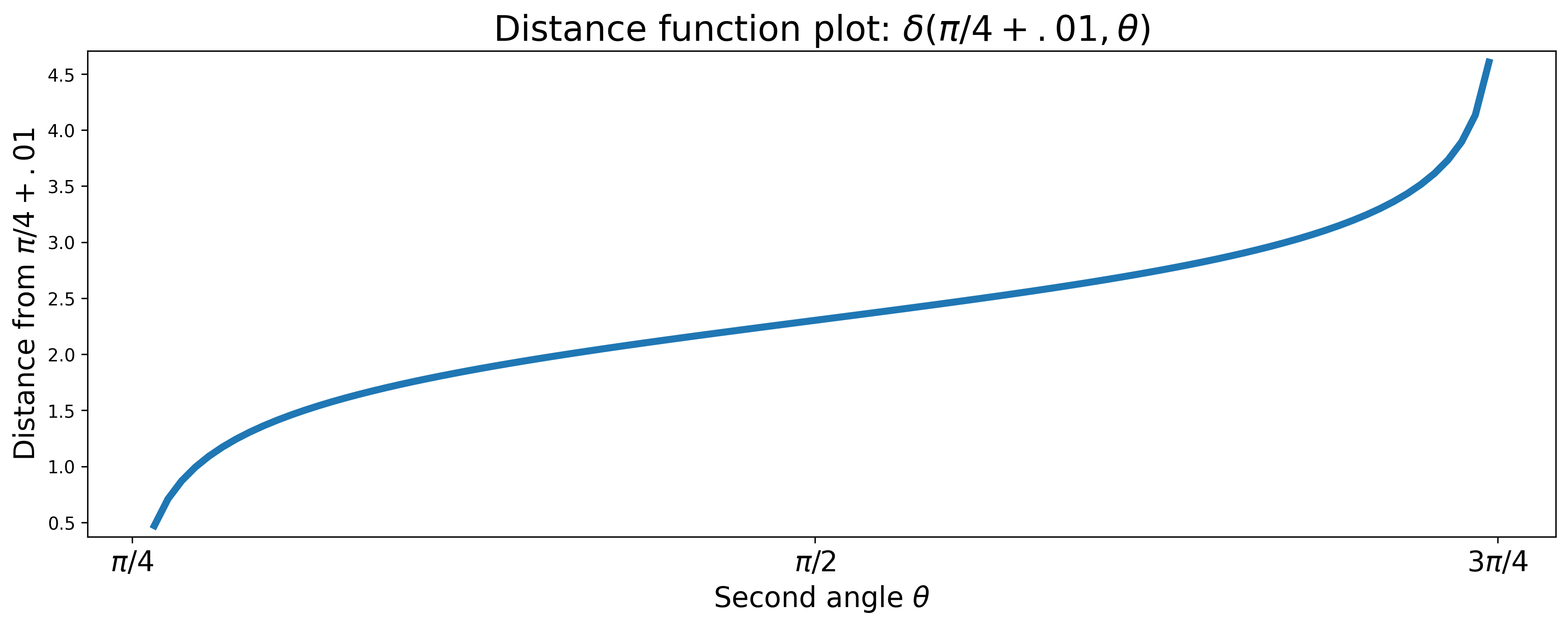}
\end{center}
\end{figure}

If we assume without loss of generality that $\sin(\theta_1) < \sin(\theta_2) < \pi/2$ (i.e. $\theta_2$ hits higher on the hyperboloid than $\theta_1$), then $\theta_{m, \text{naive}}$ will hit closer to $\theta_1$. Instead, we want some function $F(\theta_1, \theta_2) = \theta_m \in [\theta_1, \theta_2]$ such that $\delta(\theta_1, \theta_m) = \delta(\theta_2, \theta_m)$. To do this, we need to compute hyperbolic distances, so we use the pseudo-Euclidean metric in our ambient Minkowski space. In particular, we have the distance between two points defined as:
\begin{align}
% \delta(x_0, x_d) &= x_d^2 - x_0^2\\
% B(u, v) &= \frac{Q(u+v) - Q(u) - Q(v)}{2}\\
% &= x_Dy_D - x_0y_0\\
% d(u, v) &= \text{arcosh}(-B(u,v))\\
\delta(u, v) &= \cosh^{-1}(-\langle u, v \rangle_\mathcal{L})\\
&= \cosh^{-1}(x_0y_0 - x_dy_d)\\
&= \ln\left(x_0y_0 - x_dy_d + \sqrt{(x_0y_0 - x_dy_d)^2 - 1}\right)
\end{align}
This assumes that all dimensions besides $0$ and $d$ are 0, and fixes the point on the intersection between $\hdk$ and the decision hyperplane as the frame of reference for all distances. Using the definition of $\alpha(\theta, K)$ in Equation~\ref{eq:alpha}, we can simplify the conditions under which $\theta_m$ is an equidistant midpoint of $\theta_1$ and $\theta_2$:
\begin{equation}
    \delta(\theta_a, \theta_b) := \cosh^{-1}\left( \alpha(\theta_a, K)\alpha(\theta_b, K)\cos(\theta_a + \theta_b) \right)
\end{equation}
This distance function is quite nonlinear, as seen in Figure \ref{fig:angle_distances}, which corroborates the inappropriateness of simply taking the mean between two angles as a midpoint. Instead, we set the distances $\delta(\theta_1, \theta_m)$ and $\delta(\theta_m, \theta_2)$ equal and simplify. For conciseness, we define the shorthand $\alpha_n := \alpha(\theta_n, K)$:
\begin{align}
    \cosh^{-1}(\alpha_1\alpha_m\cos(\theta_1 + \theta_m)) 
    &= \cosh^{-1}(\alpha_m\alpha_2\cos(\theta_m + \theta_2))\\
    \alpha_1\alpha_m\cos(\theta_1 + \theta_m) &= \alpha_m\alpha_2\cos(\theta_m + \theta_2)\\
    \alpha_1\cos(\theta_1 + \theta_m) &= \alpha_2\cos(\theta_m + \theta_2)\\
    \frac{\sqrt{-\sec(2\theta_1)}}{\sqrt{K}}\cos(\theta_1 + \theta_m) &= \frac{\sqrt{-\sec(2\theta_2)}}{\sqrt{K}}\cos(\theta_m + \theta_2)\\
    \sec(2\theta_1)\cos^2(\theta_1 + \theta_m) &= \sec(2\theta_2)\cos^2(\theta_m + \theta_2)\\
    \frac{\cos^2(\theta_1 + \theta_m)}{\cos(2\theta_1)} &= \frac{\cos^2(\theta_m + \theta_2)}{\cos(2\theta_2)}\\
    \cos(2\theta_2)\cos^2(\theta_1 + \theta_m) &= \cos(2\theta_1)\cos^2(\theta_2 + \theta_m) \\
    \cos(2\theta_2)(\cos(\theta_1)\cos(\theta_m)\!-\! \sin(\theta_1)\sin(\theta_m))^2 
    &= 
    \cos(2\theta_1)(\cos(\theta_2)\cos(\theta_m)\!-\! \sin(\theta_2)\sin(\theta_m))^2 \\
    \cos(2\theta_2)(\cos(\theta_1)\cot(\theta_m)- \sin(\theta_1))^2 
    &= \cos(2\theta_1)(\cos(\theta_2)\cot(\theta_m)- \sin(\theta_2))^2.
\end{align}
This is a quadratic equation in $\cot(\theta_m)$
expressed as $U\cot(\theta_m)^2 + W\cot(\theta_m) + U' = 0$, where:
\begin{align}
    U &:= \cos(2\theta_2)\cos^2(\theta_1) - \cos(2\theta_1)\cos^2(\theta_2)
\nonumber \\
      &=(2\cos^2(\theta_2) - 1)\cos^2(\theta_1) - (2\cos^2(\theta_1)-1)\cos^2(\theta_2)
      \nonumber \\
    &= \cos^2(\theta_2) - \cos^2(\theta_1)\\
    W &:= \cos(2\theta_2)\cdot 2\cos(\theta_1)\sin(\theta_1) - \cos(2\theta_1)\cdot 2\cos(\theta_2)\sin(\theta_2)
    \nonumber \\
    &= \cos(2\theta_2)\sin(2\theta_1) - \cos(2\theta_1)\sin(2\theta_2) \nonumber \\
    &= \sin(2\theta_1 - 2\theta_2) \\   
    U' &:= \cos(2\theta_2)\sin^2(\theta_1) - \cos(2\theta_1)\sin^2(\theta_2)
    \nonumber\\
    &= (1-2 \sin^2(\theta_2))\sin^2(\theta_1) - (1-2\sin^2(\theta_1))\sin^2(\theta_2)
    \nonumber\\
    &= \sin^2(\theta_1) - \sin^2(\theta_2)
    \nonumber\\
    &= \cos^2(\theta_2) - \cos^2(\theta_1),
\end{align}
 Since $U=U'$ we simplify further and solve 
$\cot^2(\theta_m) - 2V\cot(\theta_m) + 1 = 0$ where:
\begin{align}
V := -W/2U &= \frac{-\sin(2\theta_1 - 2\theta_2)}{2(\cos^2(\theta_2) - \cos^2(\theta_1)})
\nonumber\\
&= \frac{-\sin(2\theta_1 - 2\theta_2)}{2(\cos(\theta_2) + \cos(\theta_1))(\cos(\theta_2) - \cos(\theta_1))}
\nonumber\\
&= \frac{-\sin(2\theta_1 - 2\theta_2)}{-8\cos(\frac{\theta_1 + \theta_2}{2})\cos(\frac{\theta_2 - \theta_1}{2})\sin(\frac{\theta_1 + \theta_2}{2})\sin(\frac{\theta_2 - \theta_1}{2})}
\nonumber\\
&= \frac{\sin(2\theta_1 - 2\theta_2)}{2\sin(\theta_2 + \theta_1)\sin(\theta_2 - \theta_1)}
\end{align}
The solutions for the quadratic equation are 
$\cot(\theta_m) = V \pm \sqrt{V^2 - 1}$; specifically, we have
\begin{equation}
\theta_m = \begin{cases}
\theta_1 & \text{ if } \theta_1 = \theta_2\\
\cot^{-1}\left( V - \sqrt{V^2 - 1} \right) & \text{ if } \theta_1 < \pi - \theta_2\\
\cot^{-1}\left( V + \sqrt{V^2 - 1} \right) & \text{ if } \theta_1 > \pi - \theta_2\\
\end{cases}
\end{equation}

\subsection{Mixture of Gaussians on hyperbolic manifolds}
\label{appendix:gaussians}
% \begin{outline}
%     \1 Geomstats for implementations
%     \1 Sample normal in $n$ dimensions
%         \2 Prepend 0 in timelike dimension
%     \1 Parallel transport
%     \1 Exponential map\footnote{May get away with just citing the paper: \cite{nagano_wrapped_2019}}\footnote{Otherwise, since we don't define parallel transport or exponential map or tangent planes earlier in the paper, we would need to do it here
%     }
% \end{outline}
We modify the method put forward in \citet{nagano_wrapped_2019} to sample Gaussians in hyperbolic space.
We briefly reiterate their method to sample a single Gaussian in $\hdk$:
\begin{enumerate}
    \item Choose a point $\mu$ in $\hdk$ to be the mean of your Gaussian.
    \item Sample $\mathbf{X}$ as $n$ samples from a Euclidean multivariate Gaussian with mean $0$ and any covariance $\mathbf{\Sigma}$ in $D$ dimensions.
    \item Transform $\mathbf{X}$ into $\mathbf{X}'$, a set of vectors in $T_{0}\hdk$ (the tangent plane of $\hdk$ at the origin), by appending 0 in the timelike dimension.
    \item Use parallel transport from the origin to $\mu$, turning $\mathbf{X}'$ into $\mathbf{X}''$, a set of vectors in $T_\mu\hdk$.
    \item Use the exponential map at $\mu$ to map $\mathbf{X}''$ to $\mathbf{X}'''$, a set of points on the surface of $\hdk$.
\end{enumerate}
For each Gaussian in our mixture, we perform this exact procedure. We choose our set of $n$ Gaussian means by sampling $n$ vectors from $\mathcal{N}(0, \mathbf{I})$ in the tangent plane and then exponentially mapping them directly to $\hdk$; in other words, we follow the above procedure but skip step (4) because $\mu$ is not defined yet (or, equivalently, because $\mu$ is the origin).

Additionally, each covariance matrix is generated by drawing $D$ $D$-dimensional samples, $\mathbf{C} \sim \mathcal{N}(0, \mathbf{I})$, and then letting $\mathbf{\Sigma} = \mathbf{CC}^T$. The entire covariance matrix is optionally rescaled by a user-set noise scalar $a$ and divided by $D$. That is,
\begin{equation}
    \mathbf{\Sigma} = a\frac{\mathbf{CC}^T}{D}.
\end{equation}
This procedure is repeated $n$ times to yield $n$ distinct covariance matrices.

Finally, class probabilities are determined by drawing $n$ values from $U(0,1)$ and normalizing them by their sum. Each point in a sample is assigned a class that determines its $\mu$ and $\mathbf{\Sigma}$. We implement this method using the geomstats package in Python \citep{miolane_geomstats_2018}, which supports vectorized versions of parallel transport and exponential maps with differing destinations.

% \subsection{Other experiments}

% \subsubsection{Geodesics do not cross decision boundary}

% \subsubsection{Euclidean dot product is superior to Minkowski}
\subsection{Equivalence of Minkowski and Euclidean dot-products}
\label{appendix:minkowski}
Most papers on hyperbolic geometry use Minkowski products. For instance, implementing the support vector machine objective in \citet{cho_large-margin_2018} relies on Minkowski products, which are crucial for the optimization procedure they specify.

In our case, it is sufficient to use the more intuitive Euclidean formulation, even though, in practice, Minkowski space is not equipped with Euclidean products. Intuitively, Euclidean inner products (dot-products) accurately capture whether a point is to one side of a plane or another, which is all that is needed for a decision tree classifier. However, we show further that Euclidean dot-products for \hyperdt\ decision boundaries have an interpretation in terms of Minkowski products:

The sparse Euclidean dot-product we determined in Equation \ref{sparse_dp_hyperboloid} is:
\begin{equation}
    S(x) = \text{sign}\left(\max\left(0,\ \left(\sin(\theta)x_d - \cos(\theta)x_0\right)\right)\right)
\end{equation}
And, since the Minkowski inner product is simply the Euclidean dot-product with the sign of the timelike dimension flipped, we can equivalently say
\begin{equation}
    S(x)_\text{Minkowski} = \text{sign}\left(\max\left(0,\ \left(\sin(\theta)x_d + \cos(\theta)x_0\right)\right)\right)
\end{equation}
Any $\theta$ in the Euclidean case, if substituted for $-\theta$ in the Minkowski case, will yield the same $\cos(\theta)$ and a negated $\sin(\theta)$. That is,
\begin{align}
    \sin(\theta)x_d - \cos(\theta)x_0 &= a\\
    \sin(-\theta)x_d + \cos(-\theta)x_0 &= b\\
    -\sin(\theta)x_d + \cos(\theta)x_0 &= b\\
    \sin(\theta)x_d - \cos(\theta)x_0 &= -b\\
    a &= -b
\end{align}
That is, for any angle $\theta$ yielding a particular split $S$ over a dataset $\mathbf{X}$, evaluating the split using Minkowski inner products with the angle $-\theta$ produces an equivalent split. The sets are exactly the same, but the sign of the dot-product is flipped. 
%Since specific 

% \subsection{Additional experiments}
% \subsubsection{Effect of depth and predictor number}
% \subsubsection{Random forests in other embeddings}

% \subsubsection{Hyperbolic Image Embeddings}

\subsection{Additional experiments}

\subsubsection{Scaling}
\label{appendix:scaling}
In Figure \ref{fig:runtimes}, we showed that \hyperdt\ runtime scales linearly with sample size---an improvement over the exponential scaling of \textsc{HoroRF}. In this section, we show how runtime scales with the number of data points, number of dimensions, maximum depth of decision trees, and total number of estimators. Since we are not comparing to slower methods, we can extend our analysis to substantially more than the upper limit of 800 samples we use in the main section of the paper.

\paragraph{Procedure.} 
Unless noted otherwise, we test the runtime and F1-micro accuracy of \hyperrf\ with a maximum depth of 3 and 12 trees, consistent with the \hyperrf\ results in Table \ref{tab:accuracies}. We restrict ourselves to Gaussian mixtures of five classes rather than two, since this more challenging classification task has a greater range of F1-micro scores. Unless noted otherwise, we generated 1,000 points for 20 distinct trials, without cross-validation and using a test set size of 200 points. 

We carried out four distinct scaling experiments, recording runtime and accuracy as we varied:
\begin{enumerate}
    \item The number of points generated from 100 to 3,000
    \item The number of dimensions from 2 to 64
    \item The total number of decision trees in a forest from 1 to 30
    \item The max depth of each decision tree from 1 to 20
\end{enumerate}

Additionally, for the final max depth experiment, we also tested \sklrn\  Euclidean random forests and \textsc{HoroRF}, also using 12 predictors. For this portion of the experiment, we restrict ourselves to 800 samples from a 2-dimensional, 2-class mixture of Gaussians.

\begin{figure}[ht]
    \centering
    \includegraphics[width=\textwidth]{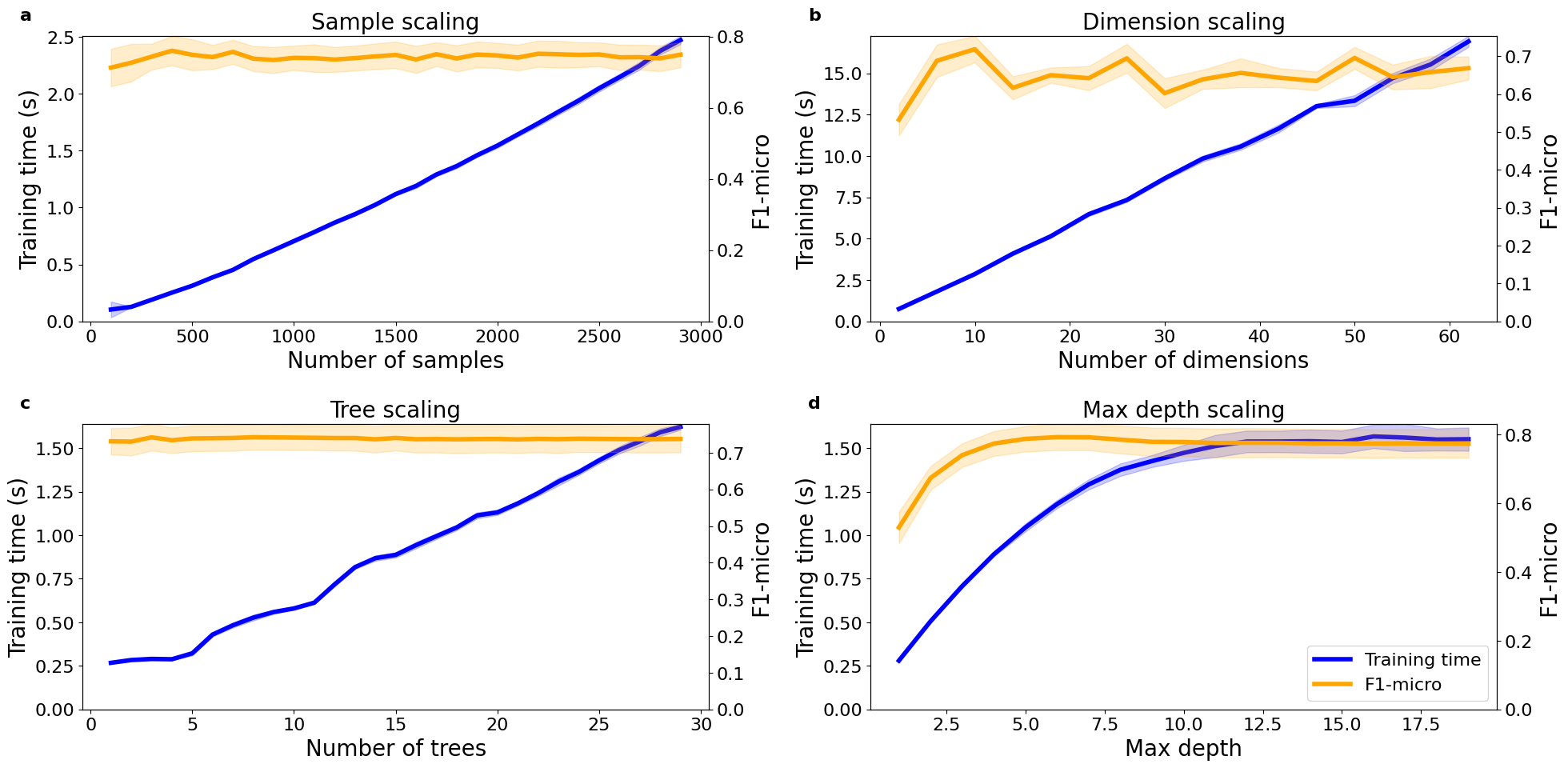}
    \caption{Observed runtimes when varying: (a) number of samples, (b) dimensionality, (c) number of trees, and (d) maximum depth in a simulated Gaussian mixture classification problem. We observe linear scaling for (a), (b), and (c), and possibly sublinear scaling with maximum tree depth. Shaded regions represent 95\% confidence intervals.}
    \label{fig:scaling1}
\end{figure}

\paragraph{Results.} 
Times and F1-micro scores for the four \hyperrf\ scaling experiments are shown in Figure \ref{fig:scaling1}. This figure shows that runtime scales linearly with the number of samples, dimensions, and trees, with little effect on overall prediction accuracy. Interestingly, the runtime levels off rather than growing exponentially for the maximum depth experiment as one might expect given the $2^{max\_depth}$ splits the predictor is allowed to make. This is because the actually achieved depth tops out when the training set is perfectly divided into homogeneous subregions of the decision space, and further splits are not made. Additionally, F1-micro scores decline slightly with increasing tree depth, likely due to overfitting.

\begin{figure}[ht]
    \centering
    \includegraphics[width=\textwidth]{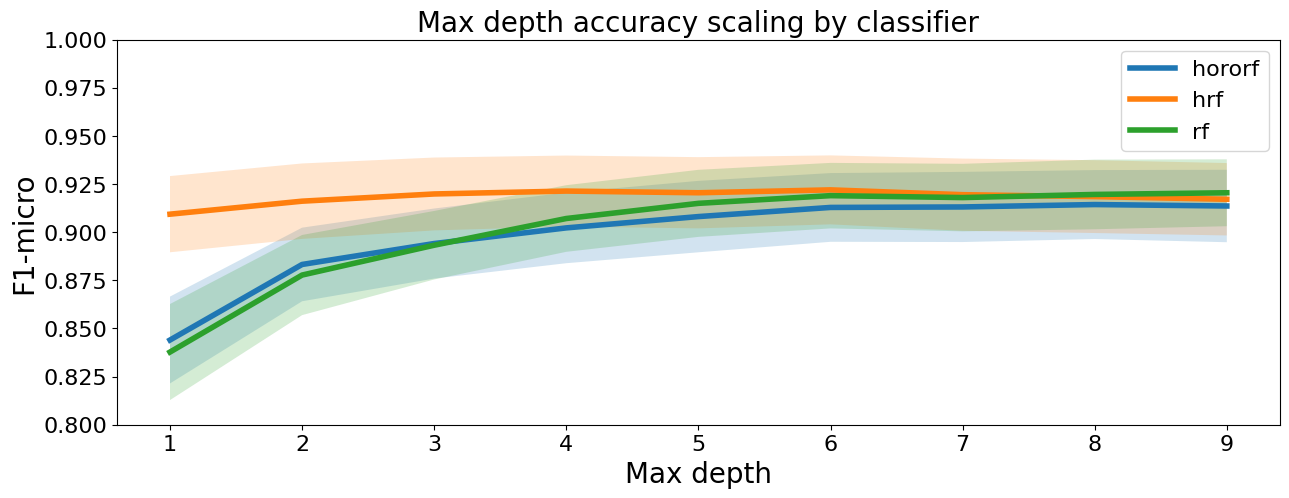}
    \caption{F1-micro scores on a Gaussian mixture classification problem with 2 classes, 2 dimensions, and 800 samples. Shaded regions represent 95\% confidence intervals.}
    \label{fig:scaling2}
\end{figure}

Since maximum depth is the only parameter with a particularly interesting relationship to prediction accuracy, we explored it further in the context of the other predictors evaluated in the paper. In Figure \ref{fig:scaling2}, we compare the F1-micro scores of \hyperrf, \hororf, and Euclidean random forests, and find that \hyperrf\ has a consistent advantage over other predictors at the same maximum depth; however, as maximum depth increases, this advantage becomes less prominent. This result speaks both to the general ability of elaborate random forests to model data from arbitrary probability distributions, and to the parsimony of \horodt-based methods in modeling hierarchical data. 

% {\textcolor{red}{[Eitan:] 
% At a depth of 1, \hyperrf outperforms Euclidean random forests by around 10\% because Hyperbolic space model hieriechal data much better than Euclidean space.
% .}}

\subsubsection{Comparison to other hyperbolic classifiers}
\label{appendix:hsvm_hmlr}
In the main body of the paper, we restricted ourselves to classifiers based on decision trees and random forests. However, there are a number of other capable and powerful classifiers for use on hyperbolic data that warrant evaluation. In this section, we evaluate hyperbolic support vector machines and logistic regression classifiers against some of our benchmarks. 

\paragraph{Procedure.}
We used the implementation of hyperbolic support vector machines provided by \citet{agibetov_using_2019}\footnote{\url{https://github.com/plumdeq/hsvm}}, and the implementation of hyperbolic logistic regression provided by \cite{bdeir_hyperbolic_2023}\footnote{\url{https://github.com/danielbinschmid/HyperbolicCV/tree/main}}. For each trial, we generated 800 points from a gaussian mixture with 2 classes and evaluated it using the F1-micro score under 5-fold cross validation. We did this for 10 trials total, in 2, 4, 8, and 16 dimensions.

In addition to the hyperbolic support vector machine and logistic regression classifiers, we evaluated their Euclidean counterparts, which are implemented in \sklrn~\cite{pedregosa_scikit-learn_2011}. We only benchmark against \hyperdt, which is simpler and slightly less accurate than \hyperrf, for fairness. 

\paragraph{Results.}
We report average F1-micro scores for each classifier in each dimension in Table \ref{tab:hsvm_mlr}. Following the conventions of the paper, we additionally mark statistically significant differences from \hyperdt\ with an asteristk. In total, \hyperdt\ was a statistically significant improvement over each classifier in at least one dimension, and was the best classifier in half the cases. This shows a consistent advantage over the other classifiers, which we can expect to be further improved by use of \hyperrf. 

\begin{table}[ht]
    \centering
    \begin{tabular}{lrrrrr}
\toprule
                    & Logistic      & Hyperbolic Logistic & Hyperbolic Support  & Support Vector & \hyperdt \\
$D$            & Regression    & Regression          & Vector Classifier   & Classifier\\
\midrule
2 & 90.11$^*$ & 88.65$^*$ & 81.50$^*$ & 80.96$^*$ & \textbf{91.88} \\
4 & 99.20 & \textbf{99.41} & 96.94$^*$ & 85.56$^*$ & 99.30 \\
8 & 99.97 & 99.96 & \textbf{100.00} & 79.06$^*$ & 99.96 \\
16 & 99.99 & \textbf{100.00} & 98.12$^*$ & 86.99$^*$ & \textbf{100.00} \\
\bottomrule
\end{tabular}

    \caption{Micro-averaged F1 scores under 5-fold cross validation averaged over 10 seeds for each classifier and dimension. Bold indicates the best score for that dimension; asterisks indicate a statistically significant difference from \hyperdt\ as determined by a paired $t$-test.}
    \label{tab:hsvm_mlr}
\end{table}

\subsubsection{Random forests in other geometries}
\label{appendix:other_geometries}
While no representation of hyperbolic space is explicitly compatible with axis-aligned splits, it is worth exploring the possibility that other representations nonetheless lend themselves better to treatment by decision tree or random forest classifiers than the hyperboloid model does; more specifically, it is worth testing \hyperdt\ and \hyperrf\ against a greater range of embeddings to ensure we are making a fair comparison.

\paragraph{Procedure.}
Analogously to other sections, we restrict ourselves to 800 samples from 2-class mixtures of Gaussians in 2, 4, 8, and 16 dimensions. We record F1-micro scores under 5-fold cross-validation, averaged over 10 seeds. In this case, we evaluated \hyperdt, \hyperrf, and scikit-learn implementations of Euclidean decision trees and random forests for each sample.

We converted each sample to Euclidean, Hyperboloid, Klein disk, and Poincar\'e disk coordinates. To get Euclidean coordinates, we applied the logarithmic map to project points from $\hdk$ to the tangent plane at the origin.

\paragraph{Results.}
Table \ref{tab:geometry_table} shows a comparison of \hyperdt\ and \hyperrf\ to each of these geometries. Both \hyperdt\ and \hyperrf\ show substantial, statistically significant advantages over their Euclidean counterparts when applied to the hyperboloid model, and this is the only consistent trend in the data. Only the Poincar\'e disk in two dimensions beat \hyperrf\ with statistical significance. 

Interestingly, the Klein disk embeddings performed well (without statistical significance) in two and three dimensionalities for decision trees and random forests, respectively. This is likely because geodesics in the Klein model are represented with straight lines, so the axis-parallel splits used by Euclidean decision tree algorithms are also geodesic decision boundaries. This is yet another point in support of geodesic decision boundaries yielding improved classification performance.

\begin{table}[h]
    \centering
    \begin{tabular}{ll|llll}
\toprule
 &  & \multicolumn{4}{c}{Dimensions} \\
Model & Geometry & 2 & 4 & 8 & 16 \\
\midrule
\multirow[t]{5}{*}{Decision Tree} & Euclidean & 91.86 & 99.15 & 99.94 & 99.97 \\
 & Hyperboloid & 90.16$^*$ & 98.34$^*$ & 99.91 & 99.99 \\
 & Klein & \textbf{91.89} & 99.29 & \textbf{99.96} & 99.99 \\
 & Poincar\'e & 91.85 & \textbf{99.30} & \textbf{99.96} & \textbf{100.00} \\
 % & Tangent PCA & \textbf{91.89} & \textbf{99.33} & \textbf{99.97} & 99.97 \\
\cline{1-6}
\hyperdt & Hyperboloid & 91.88 & \textbf{99.30} & \textbf{99.96} & \textbf{100.00} \\
\cline{1-6}
\hyperrf & Hyperboloid & 91.91 & 99.42 & 99.96 & \textbf{100.00} \\
\cline{1-6}
\multirow[t]{5}{*}{Random Forest} & Euclidean & 92.09 & 99.28 & 99.97 & \textbf{100.00} \\
 & Hyperboloid & 89.80$^*$ & 98.40$^*$ & 99.95 & \textbf{100.00} \\
 & Klein & 92.05 & \textbf{99.46} & \textbf{100.00} & \textbf{100.00} \\
 & Poincar\'e & \textbf{92.26}$^*$ & 99.45 & 99.97 & \textbf{100.00} \\
 % & Tangent PCA & \textbf{92.42}$^*$ & 99.41 & 99.97 & \textbf{100.00} \\
\bottomrule
\end{tabular}

    \caption{F1-micro scores for Euclidean and hyperbolic random forests and decision trees when applied to a variety of hyperbolic coordinate systems. Bolded scores are the best for that dimension; asterisks represent statistical significance.}
    \label{tab:geometry_table}
\end{table}

\subsubsection{Hyperbolic image embeddings}
\label{appendix:image_embeddings}
CLIP is a contrastive deep learning model with a joint text-image embedding space \citep{radford_learning_2021}. Recently, \citet{desai_hyperbolic_2023} developed MERU, a modified version of CLIP which encodes images and text into $\mathbb{H}^{512, 0.1}$. They impose a hierarchical structure on image an text embeddings: since images are more specific than the sentences which describe them, they encourage image embeddings to have larger values in dimension $0$ than their corresponding text embeddings. In this case, the root of the hyperboloid represents the most general possible concept. Learning this hierarchy, MERU matches or outperforms CLIP on zero-shot text-to-image and image-to-text \textit{retrieval} tasks. However, it matches or underperforms CLIP on zero-shot image \textit{classification} tasks. 

\paragraph{Procedure.}

We hypothesized that MERU classification performance suffers because \citet{desai_hyperbolic_2023} used a logistic regression classifier, designed for Euclidean, not hyperbolic, space. To test this, we perform zero-shot image classification on CIFAR-10 on pretrained ViT S/16 MERU and CLIP image embeddings using Euclidean and hyperbolic random forests.\footnote{Pretrained models at \url{https://github.com/facebookresearch/meru}.} 
All forests used 10 estimators with a maximum depth of 5.

We also experiment with partial and full image embeddings. CLIP passes an image throught an image-encoder, a linear projection layer, and then an $L_2$ normalization layer (project to the unit hypersphere). Similarly, MERU passes an image through an image-encoder, a linear projection layer, and then an exponential map (project to the hyperboloid). In their experiments, \citet{desai_hyperbolic_2023} performed classification only on the image-encoder outputs, which lie in Euclidean space for both CLIP and MERU. However, this approach ignores (1) the information provided by the projection layer and cannot leverage (2) the hierarchical structure gained from projecting to the hyperboloid. We thus exeriment with embeddings with different combinations of layers.

\paragraph{Results.}
The results of this experiment are summarized in Table \ref{tab:meru}, alongside the reported accuracies from the original paper. We do not report standard deviation because we use the same train/test split as \citet{desai_hyperbolic_2023}. 

First, we find that Euclidean random forests perform better on MERU-encoded data than CLIP-encoded data. This suggests that, in a linear probing context, MERU representations actually are more separable with respect to CIFAR-10 classes.

Additionally, we show that hyperboloid random forests on MERU encodings with linear projection and exponential mapping to the hyperboloid outperform all other combinations of classifiers and embeddings.
These findings both demonstrate that hyperbolic embeddings for image-text joint embeddings enhance zero-shot classification performance and that hyperbolic random forests outperform their Euclidean counterparts in this embedding space.

\hyperrf\ outperforms all Euclidean random forests, suggesting it is the best forest-type classifier for this task. However, we fail to beat the benchmark value reported in \citet{desai_hyperbolic_2023} for logistic regression. We reproduce similar accuracies on our own implementation of logistic regression.

\begin{table}[t]
    \begin{center}
    %\scalebox{0.7}{
    % \begin{tabular}{l|lllll}
% \hline
% Encoder             & MERU      & MERU      & CLIP      & CLIP      & CLIP\\
% \hline
% Classifier          & fastH     & sklearn   & sklearn   & sklearn   & sklearn\\
% Linear projection   & $\checkmark$       & $\checkmark$       & No        & $\checkmark$       & $\checkmark$\\
% Exp map/Norm        & $\checkmark$       & $\checkmark$       & No        & No        & $\checkmark$\\
% Accuracy            & \textbf{86.20\%}   & 85.05\%   & 82.00\%   & 84.20\%   & 84.00\%
% \end{tabular}

% \begin{tabular}{lll|lll}
%     \hline
%     Encoder & Linear projection & Exp map/norm  & fastH & sklearn & Original\\
%     \hline
%     MERU    & $\ding{55}$                  &               &                & 86.10\%          & 68.7\%\\
%     MERU    & $\checkmark$               & $\checkmark$           & \textbf{86.20\% }       & 85.05\% & \\
%     CLIP    &                   &               &                & 82.00\%          & 72.00\%\\
%     CLIP    &                &               &                & 84.20\%          & \\
%     CLIP    & $\checkmark$               & $\checkmark$           &                & 84.00\%          & \\
%     \hline 
% \end{tabular}

\begin{tabular}{llll}
    \hline
    Embedding               & Predictor                         & CLIP Accuracy    & MERU Accuracy\\
    \hline
    % Encoder                 & Baseline                          & 72.00            & 68.70\\
    Encoder                 & Baseline                          & 89.60            & 89.70\\
    Encoder                 & Logistic regression               & 87.60            & \textbf{90.15}\\
    Encoder                 & Random forest                     & 82.00            & 86.05\\
    Encoder + LP            & Random forest                     & 84.20            & 84.85\\
    Encoder + LP + map      & Random forest                     & 84.00            & 85.10\\
    Encoder + LP + map      & \hyperrf                          & ---              & 86.20\\
    \hline
\end{tabular}
   
    %}
    \end{center}
    \caption{Per-class average accuracies on zero-shot CIFAR-10 classification benchmarks. \hyperrf\ and \sklrn\ benchmarks are new, with baseline accuracies taken from Table 7 of \citet{desai_hyperbolic_2023}.
%    \footnote{ET: I don't think this is correct. Looking at the MERU paper, the accuracy of ViT S/16 on Cifar 10 is  89.6 on CLIP and 89.7. This is better than any of our results here! This is a big issue. Maybe we can attribute this to hyperparameter search that the authors of MERU did but that we did not do.}
    {\em LP} stands for linear projection. Note that {\em map} means $L_2$ normalization for CLIP and exponential map to the hyperboloid for MERU. \hyperrf\ can only be evaluated on MERU with linear projection and exponential map applied, since all other representations are Euclidean.}
    \label{tab:meru}
\end{table}

\subsubsection{WordNet embeddings}
\label{appendix:wordnet}
Hyperbolic embeddings of Wordnet~\cite{fellbaum_wordnet_2010} are another popular benchmark for hyperbolic classifiers. We extend our analysis to WordNet classification tasks.

\paragraph{Procedure.}
We use the WordNet embeddings and labels provided in the Github repository for \citet{doorenbos_hyperbolic_2023}\footnote{\url{https://github.com/LarsDoorenbos/HoroRF}}. These are split into binary classification tasks, where embeddings are labeled according to whether or not they belong to a certain class of things (animal, group, mammal, and so on), and multiclass classification tasks. For speed, we downsample the WordNet embeddings to 1,000 randomly-sampled points without rebalancing the classes. The exact same sample is seen by all classifiers. Each classifier is evaluated across 10 seeds using 5-fold cross-validation.

\paragraph{Results.}
The results for the WordNet experiment are shown in Table \ref{tab:wordnet}. These results continue to be very favorable for \hyperdt\ and \hyperrf: \hyperdt\ beats Sklearn and \hororf\ 8 times each, and \hyperrf beats Sklearn 6 times and \hororf\ 8 times, all with statistical significance. No other models achieved a statistically significant advantage against any other models. 

\begin{table}[t]
    \begin{center}
    \begin{tabular}{ll|lll|lll}
\toprule
 &  & \multicolumn{3}{c|}{Decision Trees} & \multicolumn{3}{c}{Random Forests} \\
 & & & \textsc{scikit-} & & & \textsc{scikit-} & \\
Data &  & \hyperdt & \textsc{learn} & \hororf & \hyperrf & \textsc{learn} & \hororf \\
%Data & $d$ & $n$ &  &  &  &  &  &  \\
\midrule
\multirow[t]{8}{*}{
\begin{sideways}
\hspace{-2cm}
Binary
\end{sideways}
}
 & Animal & \textbf{98.88}\textsuperscript{$\dagger$} & 98.69 & 96.02 & \textbf{98.97}\textsuperscript{$\ddagger$}\textsuperscript{$\dagger$} & 98.16 & 96.22 \\
 & Group & \textbf{94.65}\textsuperscript{$\ddagger$}\textsuperscript{$\dagger$} & 94.06 & 91.77 & \textbf{94.64}\textsuperscript{$\dagger$} & 94.23 & 92.34 \\
 & Mammal & \textbf{99.86}\textsuperscript{$\ddagger$}\textsuperscript{$\dagger$} & 99.33 & 98.92 & \textbf{99.87}\textsuperscript{$\ddagger$}\textsuperscript{$\dagger$} & 99.19 & 98.92 \\
 & Occupation & 99.58 & 99.49 & \textbf{99.64} & 99.61 & 99.66 & \textbf{99.69} \\
 & Rodent & \textbf{99.83} & 99.78 & 99.79 & 99.81 & \textbf{99.85} & \textbf{99.85} \\
 & Solid & \textbf{99.11}\textsuperscript{$\ddagger$}\textsuperscript{$\dagger$} & 98.72 & 98.55 & \textbf{99.13}\textsuperscript{$\ddagger$}\textsuperscript{$\dagger$} & 98.50 & 98.45 \\
 & Tree & \textbf{98.90}\textsuperscript{$\ddagger$}\textsuperscript{$\dagger$} & 98.59 & 98.46 & \textbf{99.01}\textsuperscript{$\ddagger$}\textsuperscript{$\dagger$} & 98.68 & 98.63 \\
 & Worker & \textbf{98.69}\textsuperscript{$\ddagger$} & 98.36 & 98.57 & \textbf{98.73} & 98.58 & 98.57 \\

\cline{1-8}

\multirow[t]{3}{*}{
\begin{sideways}
\hspace{-0.8cm}
Multi
\end{sideways}
} & Same level & \textbf{98.10}\textsuperscript{$\ddagger$}\textsuperscript{$\dagger$} & 97.33 & 96.98 & \textbf{98.31}\textsuperscript{$\ddagger$}\textsuperscript{$\dagger$} & 96.91 & 96.71 \\
 & Nested & \textbf{89.44}\textsuperscript{$\ddagger$}\textsuperscript{$\dagger$} & 87.74 & 77.19 & \textbf{89.72}\textsuperscript{$\dagger$} & 89.22 & 86.36 \\
 & Both & \textbf{96.38}\textsuperscript{$\ddagger$}\textsuperscript{$\dagger$} & 95.60 & 91.22 & \textbf{96.67}\textsuperscript{$\ddagger$}\textsuperscript{$\dagger$} & 94.33 & 91.13 \\

\bottomrule
\end{tabular}

    \end{center}
    \caption{Mean micro-F1 scores for classification benchmarks over 10 seeds and 5 folds. The highest-scoring decision tree and random forests are bolded separately. $*$ means a predictor beat \hyperrf, $\dagger$ means a predictor beat $\hororf$, and $\ddagger$ means a predictor beat \sklrn, with $p < 0.05$.}
    \label{tab:wordnet}
\end{table}

\subsubsection{Midpoint ablation}
\label{appendix:ablation}
Because the use of the midpoint formula laid out in Equation \ref{eq:midpoint} was guided by intuition rather than actual performance, a worthwhile experiment is to check the effect that substituting this operation with the naive midpoint calculation has on the accuracy and efficiency of our algorithm.

\paragraph{Procedure.}
We test the effect of substituting the geodesic midpoint calculations with naive midpoint calculations across 10 seeds and a single train-test split. We check this for 100, 200, 400, and 800 points and 2, 4, 8, and 16 dimensions, recording total runtime and F1-micro score on the held-out test set (20\% of the sample).

\begin{figure}
    \centering
    \includegraphics[width=\textwidth]{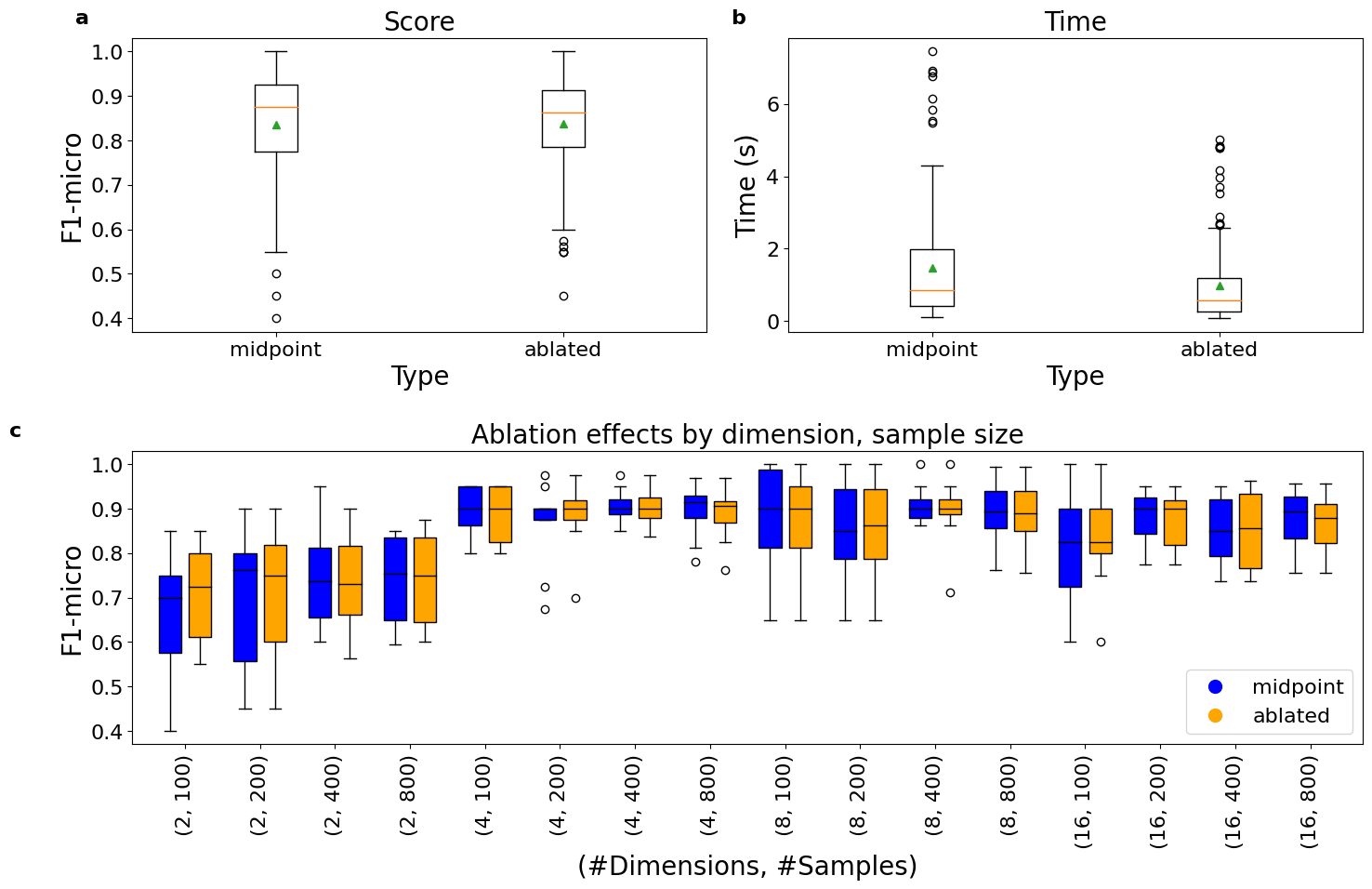}
    \caption{Ablation results for midpoint angle computations.}
    \label{fig:ablations}
\end{figure}

\paragraph{Results.}
Figure \ref{fig:ablations} shows the results of the midpoint ablation study. Across almost all sample sizes and dimensions, substituting the naive midpoint computation resulted in a marked reduction in accuracy---although with the benefit of a slight reduction in runtime. We believe that not only is this accuracy-performance tradeoff favorable to the more complicated midpoint computation, it is also theoretically more justified (all things being equal, you should prefer to place the decision boundary exactly in between two differing classes). Thus, we elect to keep the midpoint computation as it is.

% We also perform report different accuracies than \citet{desai_hyperbolic_2023} because we do not perform hyperparmater tuning.

% Initially, it seems that learning a text-image hierarchy causes MERU's image classification performance to suffer. However, we show this is not the case. We hypothesize that MERU's underperformance arises because the classifer selected, the logistic regression classifer, is designed for Euclidean space, not hyperbolic space. In fact we demonstrate it is possible to learn a highly interpretable representation space while still maintaing strong classification performance. 

\subsection{Statistical testing}
\label{appendix:p_vals}
For the benchmarks reported in Table \ref{tab:accuracies}, we provide statistical significance annotations. To test statistical significance, we used a two-tailed paired $t$-test comparing $F_1$ scores over 10 seeds and 5 folds. Each combination of dataset, dimension, and sample size was tested separately. We used a threshold of $p=0.05$ on this test to determine statistical significance, and assigned the significance annotation to the predictor with the higher mean. We report full $p$-values in Table \ref{tab:p_vals}.

\begin{table}[hb]
    \centering
    \begin{tabular}{lll|lll|lll}
\toprule
&&&  DT vs  & DT vs & HoroRF vs  & RF vs & RF vs  & HoroRF vs  \\
Dataset & $D$ & $n$ & \hyperdt & HoroRF & \hyperdt & \hyperrf & HoroRF & \hyperrf \\
\midrule
\multirow[t]{16}{*}{
\begin{sideways}
\hspace{-3.5cm}
Gaussian
\end{sideways}
} & \multirow[t]{4}{*}{2} & 100 & .204 & 4.56e-03 & .027 & .368 & 3.92e-03 & 1.16e-03 \\
 &  & 200 & .545 & .053 & 5.88e-04 & .960 & 1.39e-04 & .092 \\
 &  & 400 & 4.75e-03 & 4.86e-03 & 6.10e-05 & .867 & 6.71e-07 & 4.41e-05 \\
 &  & 800 & 1.52e-03 & 3.02e-04 & 1.05e-06 & .860 & 1.14e-10 & 2.55e-07 \\
\cline{2-9}
 & \multirow[t]{4}{*}{4} & 100 & .067 & .358 & 8.29e-05 & 1.00 & 9.91e-06 & .417 \\
 &  & 200 & .036 & 6.50e-03 & 3.98e-05 & .705 & 1.58e-07 & .025 \\
 &  & 400 & 7.02e-03 & 5.43e-03 & 5.95e-03 & .514 & 7.62e-07 & 4.90e-03 \\
 &  & 800 & 4.29e-03 & 6.03e-04 & .034 & .062 & 5.19e-05 & 2.80e-03 \\
\cline{2-9}
 & \multirow[t]{4}{*}{8} & 100 & .709 & .420 & 2.17e-03 & .252 & .233 & .083 \\
 &  & 200 & .821 & .766 & 1.14e-03 & .766 & 9.18e-04 & 1.00 \\
 &  & 400 & .785 & .569 & .012 & .533 & .011 & 1.00 \\
 &  & 800 & .133 & .103 & 2.41e-03 & .598 & 3.83e-04 & .598 \\
\cline{2-9}
 & \multirow[t]{4}{*}{16} & 100 & .261 & .420 & .146 & 1.00 & .032 & .420 \\
 &  & 200 & .322 & .569 & .011 & .261 & .028 & .322 \\
 &  & 400 & .322 & --- & .182 & .159 & .044 & .159 \\
 &  & 800 & --- & --- & .033 & .096 & --- & --- \\
\cline{1-9} \cline{2-9}
\multirow[t]{16}{*}{
\begin{sideways}
\hspace{-3.5cm}
NeuroSEED
\end{sideways}
}  & \multirow[t]{4}{*}{2} & 100 & .236 & .160 & .014 & .577 & 2.37e-03 & .824 \\
 &  & 200 & .037 & 8.32e-03 & 1.14e-05 & .138 & 1.54e-06 & .912 \\
 &  & 400 & .010 & 8.82e-06 & 6.84e-10 & .894 & 9.78e-11 & .048 \\
 &  & 800 & .878 & 1.25e-08 & 1.77e-10 & .807 & 2.86e-10 & 4.56e-03 \\
\cline{2-9}
 & \multirow[t]{4}{*}{4} & 100 & .709 & .055 & 3.45e-17 & 3.34e-05 & 3.30e-17 & 2.02e-07 \\
 &  & 200 & .322 & 7.99e-05 & 2.57e-24 & 1.61e-11 & 3.05e-24 & 9.19e-15 \\
 &  & 400 & .159 & 2.75e-05 & 2.53e-24 & 6.10e-21 & 2.86e-24 & 6.94e-23 \\
 &  & 800 & .322 & 2.00e-10 & 1.12e-26 & 5.07e-20 & 1.22e-26 & 2.71e-24 \\
\cline{2-9}
 & \multirow[t]{4}{*}{8} & 100 & .533 & 7.92e-03 & 2.65e-14 & 3.44e-10 & 2.47e-14 & 1.60e-05 \\
 &  & 200 & .622 & 6.27e-04 & 3.62e-20 & 8.01e-20 & 1.82e-19 & 4.50e-15 \\
 &  & 400 & .229 & 5.28e-08 & 4.62e-26 & 6.56e-32 & 8.02e-26 & 4.90e-27 \\
 &  & 800 & 1.00 & 6.99e-13 & 7.88e-37 & 9.02e-36 & 6.65e-37 & 1.65e-33 \\
\cline{2-9}
 & \multirow[t]{4}{*}{16} & 100 & .322 & .595 & 2.98e-07 & .061 & 3.40e-08 & .123 \\
 &  & 200 & .279 & .880 & 4.98e-07 & .213 & 2.02e-07 & .180 \\
 &  & 400 & .569 & 2.00e-03 & 6.56e-08 & .120 & 4.82e-08 & 1.47e-04 \\
 &  & 800 & .659 & 1.26e-03 & 1.40e-14 & .114 & 1.62e-14 & 2.57e-07 \\
\cline{1-9} \cline{2-9}
\multirow[t]{4}{*}{
\begin{sideways}
\hspace{-1.2cm}
Polblogs
\end{sideways}
} & 2 & 979 & .118 & .571 & 1.23e-05 & 2.38e-05 & 9.15e-07 & 4.50e-06 \\
% \cline{2-9}
 & 4 & 979 & .160 & .696 & 2.05e-07 & 1.68e-03 & 6.26e-08 & 1.63e-03 \\
% \cline{2-9}
 & 8 & 979 & .047 & .301 & 1.11e-12 & 2.48e-10 & 1.87e-11 & 6.43e-10 \\
% \cline{2-9}
 & 16 & 979 & .987 & .302 & 2.34e-10 & 3.64e-09 & 8.29e-11 & 1.65e-07 \\

\end{tabular}

    \caption{Paired $t$-test values for benchmarks over 5 folds and 10 seeds. Missing values indicate that two sets of $F_1$ scores were identical. Statistically significant values are used to generate cell annotations in Table \ref{tab:accuracies}.}
    \label{tab:p_vals}
\end{table}

\end{document}